%% file: main.tex
\documentclass[twoside,11pt]{article}

\usepackage{jmlr2e}
\usepackage{amsmath}
\usepackage{mathtools}
\usepackage{setspace}

\DeclareMathOperator{\argmin}{arg\,min}
\usepackage{xcolor}

\ShortHeadings{Dropout Training is Distributionally Robust Optimal}{Blanchet et al.}
\firstpageno{1}

\input{comments}

\begin{document}

\title{Dropout Training is Distributionally Robust Optimal}

\author{\name Jos\'e Blanchet \email jose.blanchet@stanford.edu \\
       \addr Department of Management Science and Engineering\\
      Stanford University\\
       Stanford, CA 94305, USA
       \AND
       \name Yang Kang \email yangkang@stat.columbia.edu \\
       \addr Department of Statistics\\
       Columbia University\\
       New York, NY 10027, USA
       \AND Jos\'e Luis Montiel Olea \email
       montiel.olea@gmail.com\\
       \addr Department of Economics\\
       Columbia University\\
       New York, NY 10027, USA
       \AND Viet Anh Nguyen \email 
       viet-anh.nguyen@stanford.edu\\
       \addr Department of Management Science and Engineering\\
      Stanford University\\
       Stanford, CA 94305, USA
       \AND Xuhui Zhang \email
       xuhui.zhang@stanford.edu \\
       \addr Department of Management Science and Engineering\\
      Stanford University\\
       Stanford, CA 94305, USA}

\editor{}

\maketitle

\input{0abstract}

\onehalfspacing

\section{Introduction}

\input{1Intro}

\section{Dropout Training in Generalized Linear Models} \label{sect:glm}

\input{2GLM_and_Dropout}

\section{Problem Setup} \label{sec:Setup}

\input{3DRO}

\section{Dropout Training is Distributionally Robust Optimal} \label{sec:DRO_is_optimal} 

\input{4DropoutDRO}

\section{Statistical Guidance on Choosing $\delta$}
\label{sec:Choose_delta}

\input{9New_Section_choicedelta}

\section{An Algorithm for Dropout Training}
\label{sec:unbias}

\input{5Algorithm}


\section{Numerical Experiment}
\label{sec:numerics}

\input{6Numerics}

\section{Extensions} \label{sec:Extensions}

\input{7Extensions}

\section{Concluding Remarks} \label{sec:Conclusion}

\input{8Conclusions}


\acks{We would like to thank Matias Cattaneo, Max Farrell, Michael Leung, Ulrich M\"uller, Mark Peletier, Hashem Pesaran, Ashesh Rambachan, Roger Moon, Frank Schorfheide, Stefan Wager, and participants at the Statistics Seminar series at Columbia University for helpful comments and suggestions.  Jos\'e Blanchet acknowledges support from NSF grants 1915967, 1820942, 1838576 and the Chinese Merchant Bank. Material in this paper is based upon work supported by the Air Force Office of Scientific Research under award number FA9550-20-1-0397.}

\newpage

\appendix

\section{}

\input{Appendix}



\newpage

\bibliography{main.bbl}

\end{document}

%% file: comments.tex
\newtheorem{assumption}{Assumption}

\newcommand{\be}{\begin{equation}}
\newcommand{\ee}{\end{equation}}
\newcommand{\bea}{\begin{equation*}\begin{aligned}}
\newcommand{\eea}{\end{aligned}\end{equation*}}

\newcommand{\R}{\mathbb{R}}

\newcommand{\wh}{\widehat}
\newcommand{\mc}{\mathcal}
\newcommand{\mbb}{\mathbb}

\newcommand{\PP}{\mbb P}
\newcommand{\Pnom}{\wh{\mbb P}}
\newcommand{\QQ}{\mbb Q}

\newcommand{\opt}{^\star}
\newcommand{\eps}{\varepsilon}

\newcommand{\EE}{\mathbb{E}}

\newcommand{\half}{\frac{1}{2}}

%% file: 0abstract.tex
\begin{abstract}
This paper shows that dropout training in Generalized Linear Models is the minimax solution of a two-player, zero-sum game where an adversarial nature corrupts a statistician's covariates  using  a  multiplicative nonparametric  errors-in-variables  model. In this game, nature's \emph{least favorable distribution} is \emph{dropout noise}, where nature independently deletes entries of the covariate vector with some fixed probability $\delta$. This result implies that dropout training indeed provides out-of-sample expected loss guarantees for distributions that arise from multiplicative perturbations of in-sample data. In addition to the decision-theoretic analysis, the paper makes two more  contributions. First, there is a concrete recommendation on how to select the tuning parameter $\delta$ to guarantee that, as the sample size grows large, the in-sample loss after dropout training exceeds the true population loss with some pre-specified probability. Second, the paper provides a novel, parallelizable, Unbiased Multi-Level Monte Carlo algorithm to speed-up the implementation of dropout training. Our algorithm has a much smaller computational cost compared to the naive implementation of dropout,  provided the number of data points is much smaller than the dimension of the covariate vector.\\

\small	
 \noindent {\textit{Keywords:}} Generalized Linear Models, Distributionally Robust Optimization, Machine Learning, Minimax Theorem, Multi-Level Monte Carlo. 
\end{abstract}

%% file: 1Intro.tex
\emph{Dropout training} is an increasingly popular estimation method in machine learning.\footnote{Section 7.12 of \cite{DLbook:2016} provides a textbook treatment on dropout training.  \cite{Bishop:1995} and \cite{DropoutJMLR:2014} are seminal references on this topic.} The general idea consists in ignoring some dimensions of the covariate vector at random while estimating the parameters of a  statistical model. A common motivation for dropout training is that the random feature selection implicitly performs \emph{model averaging}, potentially improving out-of-sample prediction error and thus mitigating overfitting. See \cite{Hinton:2012} for a discussion about this point in the context of neural networks. See also \cite{Draper:1994} and \cite{Madigan:1997} for classical results on the optimality of model averaging for prediction purposes.

Our main goal is to contribute to the growing literature explaining the success of dropout training in mitigating overfitting; e.g.,~\cite{ref:wager2013dropout}, \cite{Hembold2015}, \cite{Wei2020}. Our main result (Theorem \ref{thm:glm}) shows that dropping out input features when training Generalized Linear Models can be viewed as the minimax solution to an adversarial game known in the stochastic optimization literature (\cite{Shapiro:2014}) as a Distributionally Robust Optimization (DRO) problem.

Broadly speaking, a DRO problem is a two-player, zero-sum game between a decision maker (a statistician) and an adversary (nature). The statistician wishes to choose an action to minimize a given expected loss (e.g., squared loss in a typical linear regression setting or, more generally, the negative of the log-likelihood function), while nature intends this loss to be maximal. We consider a framework in which nature is allowed to harm the statistician by corrupting the available data using a multiplicative nonparametric errors-in-variables model; as in the classical work of \cite{Hwang:1986}. The statistician is aware of the data corruption and knows the distribution used by nature, but does not have access to the realizations of the corruption noise. Under mild assumptions, nature's \emph{least favorable distribution} in this game is shown to be \emph{dropout noise}, where nature independently deletes entries of the covariate vector with some fixed probability $\delta$.  The Minimax Theorem \citep{VN:1953} is shown to also hold for this game: the minimax value coincides with the maximin solution, and these coincide with the payoffs in the game's Nash equilibrium. One direct consequence is that the statistician's selected procedure in the face of multiplicative nonparametric noise maintains optimal performance even if the adversary is allowed to corrupt after the statistician uses the training data.

Our main result (Theorem \ref{thm:glm}) shows that, by construction, dropout training indeed provides out-of-sample performance guarantees for distributions that arise from multiplicative perturbations of in-sample data. More precisely, given any fixed sample size, the \emph{out-of-sample expected loss} is no larger than that obtained by dropout training \emph{in-sample}, provided we consider out-of-sample distributions obtained as multiplicative perturbations of the in-sample distribution. Therefore, our result formally qualifies the ability of dropout training to enhance out-of-sample performance, which is one of the reasons often invoked to use the dropout method. Moreover, our results show that for any parameter value the loss used in dropout training is larger than the negative log-likelihood of Generalized Linear Models.  

We make two additional contributions. First, we suggest a novel procedure to select the  dropout probability $\delta$. To this end, we study how often the in-sample loss of dropout training exceeds the true (and unknown) population expected loss. When $\delta=0$, the Central Limit Theorem implies this event occurs with approximately .5 probability. When $\delta$ is fixed, Theorem \ref{thm:glm} implies this event occurs with probability 1. We show that picking $\delta$ to be of the form $c/\sqrt{n}$ (where $n$ denotes the number of training examples) makes the in-sample loss of dropout training exceed the population loss with probability that depends on $c$. Consequently, by choosing a target probability, say 95\%, it is possible to provide a concrete recommendation for the selection of $c$, and therefore, of $\delta$.  

Second, we suggest a new stochastic optimization implementation of dropout training. A well-known drawback of dropout is its computational cost. As we will explain, a $d$-dimensional covariate vector requires $2^d$ evaluations of the loss in order to integrate out the dropout noise for a particular data point. The computational cost is alleviated by implementing dropout training by using either Stochastic Gradient Descent (\cite{Robbins:1951}) or naive Monte-Carlo approximations to the expected loss, both of which require draws from the joint distribution of the data and dropout noise.  Unfortunately, both of these approximations introduce bias to the solution of dropout training. Also, none of these procedures can exploit the increasing availability of parallel computing in order to alleviate their computational burden. We borrow ideas from the Multi-level Monte Carlo literature---in particular from the work of \cite{ref:blanchet2019unbiased}---to suggest an unbiased (in a sense we will make precise) dropout training routine that is easily parallelizable and that has a much smaller computational cost compared to naive dropout training methods when the number of features is large (Theorem \ref{thm:unbias}). Our algorithm thus complements the recent literature suggesting approaches to speed-up dropout training by either using a parallelized implementation of  Stochastic Gradient Descent \citep{PSGD:2010} or a fast dropout training based on Gaussian approximations  \citep{ref:wang2013fast}.

The rest of the paper is organized as follows. Section \ref{sect:glm} explains dropout training in the context of Generalized Linear Models. Section \ref{sec:Setup} presents a general description of the DRO framework used in this paper. Section \ref{sec:DRO_is_optimal} specializes the DRO problem by using the negative log-likelihood of Generalized Linear Models to define a loss function for the statistician, and by allowing nature to harm the statistician via a multiplicative errors-in-variables model for the covariates. This section also presents our main theorem. Section \ref{sec:Choose_delta} presents our approach to select the dropout probability, $\delta$. Section \ref{sec:unbias} discusses different computational methods available for implementing dropout training (full  integration, Stochastic Gradient Descent, Naive Monte Carlo integration) and presents our suggested \emph{Unbiased Multi-level Monte Carlo} algorithm.  Section \ref{sec:numerics} presents some simulations comparing our recommended selection of $\delta$ to cross-validation, as well as our preferred implementation of dropout training to Stochastic Gradient Descent. Finally, Section \ref{sec:Extensions} discusses extensions of our results to a particular class of feed-forward neural networks with a single hidden layer. We show that dropout training of the hidden units in the hidden layer is distributionally robust optimal. All the proofs are collected in the Appendix. 

%% file: 2GLM_and_Dropout.tex
This section describes dropout training in the context of Generalized Linear Models.
As some other recent papers in the literature, we view Generalized Linear Models as a convenient, transparent, and relevant framework to better understand the theoretical and algorithmic properties of dropout training. 


\subsection{Generalized Linear Models (GLMs)}
A Generalized Linear Model---with parameters $\beta$ and $\phi$---is defined by a conditional density for the response variable $ Y \in\mathcal{Y}\subseteq\R$ given $X  \in\mathbb{R}^d$
\begin{equation} \label{eqn:GLM}
    f(Y|X, \beta,\phi) \equiv h(Y,\phi)\exp\left(\left(Y \beta^\top X-\Psi(\beta^\top X)\right)/a(\phi)\right),
\end{equation}
see~\citet[Equation~2.4]{mccullagh2018generalized}. In our notation $h(\cdot,\phi)$ is a real-valued function (integrable with respect to the true data distribution),   parameterized by $\phi$ defined on the domain $\mathcal{Y}$; $a(\cdot)$ is a positive function of $\phi$; and $\Psi(\cdot)$ is the log-partition function, which we assume to be defined on all the real line. It is well-known that in GLMs with a scalar response variable the log-partition function is infinitely differentiable and strictly convex on its domain;  see Proposition 3.1 in \cite{wainwright2008graphical}.

Normal, Logistic, and Poisson Regression have conditional densities of the form  (\ref{eqn:GLM}). For the sake of exposition, we provide details below for linear and logistic regression. 

\begin{example}[Linear regression with unknown variance]
Consider the linear model $Y = \beta^\top X + \eps$, in which $\eps \sim \mc N(0, \sigma^2)$ with unknown variance $\sigma^2 \in \R_{++}$ and $\eps \bot X$. The conditional distribution of $Y$ given $X$ satisfies~\eqref{eqn:GLM} with $\phi = \sigma^2$, $a(\phi) = \phi$, $\Psi(\beta^\top X) = (\beta^\top X)^2/2$ and $h(Y, \phi) = (2\pi\phi)^{-\frac{1}{2}} \exp(-Y^2/(2\phi))$.
\end{example}

\begin{example}[Logistic regression]
    Consider $Y | X \sim \text{Bernoulli}(1/(1+ \exp(-\beta^\top X))$ with $\mc Y = \{0, 1\}$. The conditional probability mass function of Y given X satisfies~\eqref{eqn:GLM} with $a(\phi) = 1$, $\Psi(\beta^\top X) = \log(1+\exp(\beta^\top X))$ and $h(Y, \phi) = 1$.
\end{example}

\noindent Generalized Linear Models are typically estimated via Maximum Likelihood using (\ref{eqn:GLM}).  Given $n$ i.i.d.~data realizations or \emph{training examples} $(x_i, y_i)$, the Maximum Likelihood estimator $(\widehat{\beta}_{\textrm{ML}}, \widehat{\phi}_{\textrm{ML}})$ is defined as any solution of the problem
\begin{equation} \label{eqn:MLE}
    \min_{\beta,\phi} \sum_{i=1}^{n} -\ln f( y_i | x_i, \beta, \phi).
\end{equation}
Maximum Likelihood estimators in GLMs are known to be consistent and asymptotically normal under mild regularity conditions on the joint distribution of $(X_i,Y_i)$ \citep{fahrmeir1985consistency}. We denote this distribution as $P^\star$.

\subsection{Dropout Training} An alternative to standard Maximum Likelihood estimation in GLMs is dropout training. The general idea consists in ignoring some randomly chosen dimensions of $x_i$ while training a statistical model. 

For a given covariate vector $x_i$---and an user-selected  constant, $\delta \in [0,1)$---define the $d$-dimensional random vector 
\begin{equation*}
    \xi_i = (\xi_{i,1}, \ldots, \xi_{i,d})^{\top} \in \{0, 1/(1-\delta) \}^d,
\end{equation*}
where each of the $d$ entries of $\xi_{i}$ is an independent draw from a scaled Bernoulli distribution with parameter $1-\delta$. This is, for $j=1,\ldots, d$:
\begin{equation} \label{eqn:DropoutNoise}
    \xi_{i,j} = \begin{cases}
    0 & \text{ with probability }  \delta, \\ (1-\delta)^{-1} & \text{ with probability }  (1-\delta).
    \end{cases}
\end{equation}
Note that when $\delta = 0$, the distribution of $\xi_{i,j}$ collapses to $\xi_{i,j} = 1$ with probability 1.
Let $\odot$ denote the binary operator defining element-wise multiplication between two vectors of the same dimension. Consider the covariate vector  
\begin{equation}
    x_i \odot \xi_i \equiv (x_{i,1} \xi_{i,1}, \ldots, x_{i,d} \xi_{i,d} )^{\top}.
\end{equation}
Some entries of the new covariate vector are 0 (those for which $\xi_{i,j}=0$) and the rest are equal to $x_{i,j}/(1-\delta)$. 

In a slight abuse of notation, let $\mathbb{E}_{\delta}$ denote the distribution of the random vector $\xi_i$, whose distribution is parameterized by $\delta$. The estimators of $(\beta,\phi)$ obtained by \emph{dropout training} correspond to any parameters $(\widehat{\beta}(\delta),\widehat{\phi}(\delta))$ that solve the problem
\begin{equation} \label{eqn:Dropout_objective}
     \min_{\beta,\phi}~\frac{1}{n} \sum_{i=1}^{n} \mathbb{E}_{\delta} \left[ -\ln f( y_i | x_i \odot \xi_{i}, \beta, \phi) \right].
\end{equation}


One possibility to solve (\ref{eqn:Dropout_objective}) is to use Stochastic Gradient Descent (\cite{Robbins:1951}). This is tantamount to i) taking a draw of $(x_i,y_i)$ according to its empirical distribution, ii) independently taking a draw of $\xi_i$ using the distribution in \eqref{eqn:DropoutNoise} and iii) computing the stochastic gradient descent update using 
\[
    \nabla \ln f( y_i | x_i \odot \xi_{i}, \beta, \phi).
\]
We provide further details about the Stochastic Gradient Descent implementation of dropout training in Section \ref{sec:unbias}.

\subsection{Question of Interest} Adding noise to the Maximum Likelihood objective in \eqref{eqn:MLE} seems, at first glance, arbitrary. Our first obvious observation is that dropout training estimators will generally not share the same probability limit as the Maximum Likelihood estimators whenever $\delta \neq 0$. This can be formalized under the following assumptions:
\begin{assumption} \label{assump:bounded}
The log-partition function $\Psi(\cdot)$ has a bounded second derivative.
\end{assumption}

\begin{assumption} \label{assump:moments}
The second moment matrix $\mathbb{E}_{P^\star}[X X^\top]$ is finite,  positive definite. 
\end{assumption}

\begin{proposition}[Consistency] \label{prop:limit}
Suppose that Assumptions \ref{assump:bounded} and \ref{assump:moments} hold. Then for any sequence  $\delta_n\to\delta\in[0,1)$ as $n\to\infty$,  $\widehat{\beta}(\delta_n)$ converges in probability to 
\begin{equation} \label{eqn:prob_limit}
\beta^\star (\delta) \equiv\adjustlimits\argmin_{\beta} \:  \mathbb{E}_{P^\star} \left[ \mathbb{E}_{\delta} \left[ -\ln f( Y | X \odot \xi, \beta, \phi) \right] \right],  
\end{equation}
where the minimizer in (\ref{eqn:prob_limit}) is unique and does not depend on $\phi$.
\end{proposition}

\begin{proof}
See Appendix \ref{subsection:limit}.
\end{proof}

The proof of this result consists of expressing dropout training as an extremum estimator and verifying standard conditions for consistency in \cite{newey1994large}. The main message of the proposition above is that the parameter $\beta\opt(0)$---which gives the probability limit of the Maximum Likelihood estimator---generally differs from $\beta\opt(\delta)$ when $\delta \neq 0$, which is the probability limit of the dropout estimator.\footnote{Relatedly, \cite{farrell}---who study deep neural networks and their use in semiparametric inference---report that their numerical exploration of dropout increased bias and interval length compared to nonregularized models.} To further illustrate this point, it is helpful to workout the details of the probability limit of the dropout estimator in the linear regression model. Algebra shows that in this model
\[ \beta^\star(\delta) = \left( \mathbb{E}_{P^\star}[XX^{\top}] + (\delta/1-\delta) \textrm{diag}(\mathbb{E}_{P^\star}[XX^{\top}]) \right)^{-1} \mathbb{E}_{P^\star}[YX],    \]
which can be interpreted as a population version of the Ridge estimator. This estimator differs from the best linear predictor of $y$ using $x$ as long as $E_{P^\star}[YX] \neq 0$ and $\delta \neq 0$. This estimator differs from Ridge regression in that $\mathbb{E}_{P^\star}[XX^{\top}]$ replaces the identity matrix (and this simple adjustment makes the estimator scale equivariant).

Despite the lack of consistency, there is some literature that has provided empirical evidence that using intentionally corrupted features for training has the potential to improve the performance of machine learning algorithms; see \cite{Maaten:2013}. Even if one is willing to accept that corrupting features is desirable for estimation, the choice of dropout noise in \eqref{eqn:DropoutNoise} remains quite arbitrary. 

The main contribution of this paper is to provide a novel decision-theoretic interpretation of dropout training (in the population and in the sample). We will argue there is a natural two-player, zero-sum game between a decision maker (statistician) and an adversary (nature) in which dropout training emerges naturally as a minimax solution. In this game, dropout noise turns out to be nature's least favorable distribution, and dropout training becomes the statistician's optimal action. The framework we use is known in the stochastic optimization literature as Distributionally Robust Optimization and we describe it very generally in the next section. We will therein also revisit the interpretation of $\beta^*(\delta)$ in the linear regression model. 

%% file: 3DRO.tex

Consider a general problem where there is a multivariate predictor $X\in\mathbb{R}^{d}$ and a scalar outcome variable $Y\in\mathbb{R}$. A Distributionally Robust Optimization (DRO) problem is a simultaneous two-player zero sum game between a decision maker (statistician) and an adversary (nature).\footnote{A seminal reference is the robust inventory control problem of~\cite{Scarf:1958}. 
Recent references describing the use of distributionally robust stochastic programs (as those considered in this paper) are~\cite{delage:2010} and~\cite{shapiro:2017}. \cite{christensen2019counterfactual} used distributionally robust optimization to characterize the sensitivity of counterfactual analysis with respect to distributional assumptions in a class of structural econometric models.
} In this section we describe the action space for each player, their strategies,  and the payoff function. 

{\scshape Actions and Payoff:} The statistician's action space consists of vectors $\theta \in \Theta$. The ranking of the statistician's actions is contingent on the realization of $(X,Y)$, and this is captured by a real-valued loss function $\ell(X,Y,\theta)$.
We assume that the statistician is called to choose an action before observing the realization of $(X,Y)$. If the statistician knew the distribution of $(X,Y)$---which we denote by $\QQ$---the statistician's preferred choice of $\theta$ would be the solution to
\begin{equation} \label{eqn:Stat_obj}
    \inf_{\theta \in \Theta} \EE_{\QQ} \left[ \ell(X,Y,\theta) \right].
\end{equation}

Instead of assuming that the distribution $\mathbb{Q}$ is exogenously determined, we think of the distribution $\QQ$ as being chosen by nature. Thus, nature's action space consists of a set of probability distributions denoted as $\mathcal{U}$. We refer to this set as the \emph{distributionally uncertainty set}. If nature knew the action selected by the statistician, nature's preferred action would be
\begin{equation} \label{eqn:nature_obj}
    \sup_{\QQ \in \mathcal{U}} \EE_{\QQ}[\ell(X,Y,\theta)].
\end{equation}

{\scshape Strategies and Solution:} The choice of $\theta$ and $\QQ$ are assumed to happen simultaneously. A statistician's strategy for this game consists of a choice of $\theta$. Likewise, nature's strategy for this game consists of a choice of $\QQ$.

A \emph{Nash equilibrium} for this game is a pair $(\theta\opt,\QQ\opt)$ such that: a) given $\QQ\opt$, the parameter $\theta\opt$ solves \eqref{eqn:Stat_obj} and b) given $\theta\opt$, the distribution $\QQ\opt$ solves \eqref{eqn:nature_obj}. 

The \emph{minimax solution} for this game is a pair $(\theta^*,\QQ^*)$ that solves
\begin{subequations}
\begin{equation} \label{eqn:minimax}
    \adjustlimits\inf_{\theta \in \Theta} \sup_{\QQ \in \mathcal{U}}~ \EE_{\QQ}[\ell(X,Y,\theta)],
\end{equation}
while the \emph{maximin} solution is based on the program
\begin{equation} \label{eqn:maximin}
     \adjustlimits\sup_{\QQ \in \mathcal{U}} \inf_{\theta \in \Theta}~ \EE_{\QQ}[\ell(X,Y,\theta)].
\end{equation}
\end{subequations}
If $\QQ\opt$ solves \eqref{eqn:maximin}, we say that $\QQ\opt$ is \emph{nature's least favorable distribution}. The mathematical program in \eqref{eqn:minimax} is typically referred to as a DRO problem. \\


%% file: 4DropoutDRO.tex
The previous section provided a general description of a Distributionally Robust Optimization problem. We specialize the general framework of Section \ref{sec:Setup}  by imposing two restrictions. First, we use the negative log-likelihood of Generalized Linear Models \citep{mccullagh2018generalized} as a loss function for the statistician. Second, we define nature's uncertainty set (i.e., the possible data distributions that nature can take) using the  multiplicative errors-in-variables model of \cite{Hwang:1986}.

\subsection{Statistician's Payoff} We define the loss function for the statistician to be the negative of the logarithm of the likelihood in \eqref{eqn:GLM}, that is,
\begin{equation} \label{eqn:loss_GLM}
    \ell(X,Y,\theta) =  -\ln h(Y,\phi) + (\Psi(\beta^\top X)-Y(\beta^\top X) )/a(\phi) ,
\end{equation}
where $\theta \equiv (\beta^\top,\phi^\top)^\top \in \Theta$. Equation \eqref{eqn:loss_GLM} defines the statistician's objective and its set of actions.

\subsection{Nature's Distributionally Uncertainty Set}
We now define the possible distributions that nature can choose. We start out by letting $\mathbb{Q}_0$ denote some benchmark or \emph{reference} distribution over $(X,Y)$. This distribution need not correspond to that induced by a Generalized Linear Model. In other words, our framework allows for the statistician's model to be misspecified.  

Next, we define nature's action space by considering perturbations of $\mathbb{Q}_0$. Although there are different ways of doing this---for example, by using either $f$-divergences (such as the Kullback-Leibler as in~\cite{ref:nguyen2020distributionally-3}) or the optimal transport distance (such as the Wasserstein distance as in~\cite{BKM2016}) to define a neighborhood---we herein use a nonparametric multiplicative errors-in-variables model as in \cite{Hwang:1986}. 


The idea is to allow nature to independently introduce measurement error to the  covariates  using multiplicative noise. Multiplicative errors have found different applications in empirical work across different disciplines, from economics to epidemiology. For example,   \cite{Attanasio2009estimating} use it to account for measurement error in consumption data when estimating the elasticity of intertemporal substitution via Euler equations. \cite{pierce1992errors} and \cite{lyles1997detailed} use it to relate health outcomes to the exposure of a chemical toxicant that is observed with error. Moreover, due to privacy considerations,  statistical agencies such as the U.S.~Census Bureau sometimes mask data using multiplicative noise; see the discussion in \cite{kim2003multiplicative} and \cite{nayak2011statistical}. Examples of datasets that contain variables masked with multiplicative noise include the Commodity Flow Survey Data (2017), the Survey of Business Owners (2012)---both from  the U.S.~Census Bureau---and the U.S.~Energy Information Administration Residential Energy Consumption survey. 

Let $\xi \equiv (\xi_1, \ldots, \xi_d)^{\top}$ be defined as a $d$-dimensional vector of random variables that are independent of $(X,Y)$. We perturb the distribution $\mathbb{Q}_0$ by considering the transformation
\begin{equation*} 
    (X,Y) \mapsto (X_1 \xi_1,\ldots, X_d \xi_d,Y)^{\top}. 
\end{equation*}
As a result, each covariate $X_j$ is distorted in a multiplicative fashion by $\xi_j$. We often abbreviate $(X_1 \xi_1, \ldots, X_d \xi_d)^{\top}$ by $X \odot \xi$, where $\odot$ is the element-wise multiplication. 

We restrict the distribution of $\xi$ in the following way. First, for a parameter $\delta \in [0, 1)$, we define  $\mathcal Q_j(\delta)$ to be the set of distributions for $\xi_j$ that are supported on the interval $[0, 1/(1-\delta)]$ and that have mean equal to 1. More specifically,
\begin{equation} \label{eqn:Qj}
    \mathcal{Q}_j(\delta) \equiv \left\{\QQ_j: \: \QQ_j \textrm{ is a probability distribution on $\mathbb{R}$,} ~ \QQ_j([0, (1-\delta)^{-1}]) = 1,~ \EE_{\QQ_j}[\xi_j]=1 \right\}.
\end{equation}
This set of distributions prescribed using support and first-order moment information is popular in the DRO literature thanks to its simplicity and tractability~\citep{ref:wiesemann2014distributionally}. From the perspective of an errors-in-variables model, these distributions are also attractive because they preserve the expected value of the covariates, assuming that $X_j$ and $\xi_j$ are drawn independently. 

Consider now the joint random vector $(X, Y, \xi) \in \R^d \times \R \times \R^d$. For a constant $\delta \in [0,1)$ consider the joint distributions over $(X, Y, \xi)$ defined by
\begin{equation} \label{eqn:natures_choice}
    \mathcal{U}(\mathbb{Q}_0,\delta) = \left\{ \mathbb{Q}_0 \otimes \QQ_1 \otimes \ldots \otimes \QQ_d: \QQ_j \in \mathcal{Q}_j(\delta)~\forall j = 1, \ldots, d \right\},
\end{equation}
where $\otimes$ is used to denote the product measure (meaning that the joint distribution is the product of the independent marginals $\mathbb{Q}_j$, $j=0, \ldots, d$). Thus, in the game we consider $\mathcal{U}(\mathbb{Q}_0, \delta)$ is nature's action space or \emph{nature's distributionally uncertainty set}. 

We will make only one assumption about the reference distribution $\mathbb{Q}_0:$
\begin{assumption} \label{asn:reg}
The distribution $\QQ_0$ satisfies $\EE_{\QQ}[ \ell( X\odot \xi, Y, \theta) ] < \infty$ for any $\QQ \in \mathcal{U}(\QQ_0,\delta)$, any $\theta \in \Theta$, and any scalar $\delta \in [0, 1)$.
\end{assumption}

This assumption implies a minimal regularity condition to guarantee that the expected loss is well-defined for both the statistician and nature. Assumption 3 is trivially satisfied when $\QQ_0$ is the empirical distribution of the data, which is one of the main cases of interest in the paper. 

\subsection{Dropout Training is DRO} We now present the main result of this section. 

\begin{theorem} \label{thm:glm}
Consider the two-player zero sum game where the statistician has the loss function in \eqref{eqn:loss_GLM} and nature has the action space in \eqref{eqn:natures_choice} for some reference distribution $\mathbb{Q}_0$ and a scalar $\delta \in [0, 1)$. If Assumption \ref{asn:reg} is satisfied, then the \textit{minimax solution} of the two-player zero sum game defined by \eqref{eqn:loss_GLM} and~\eqref{eqn:natures_choice} 
\begin{equation}\label{eq:dro}
    \adjustlimits\inf_{\theta \in \Theta} \sup_{\QQ \in\mathcal{U}(\mathbb{Q}_0,\delta) }\EE_{\QQ} \left[ \ell( X\odot \xi, Y, \theta)\right]
\end{equation}  
is equivalent to
\be \label{eq:dro-refor}
    \inf_{\theta \in \Theta}~ \EE_{\QQ\opt}[\ell(X \odot \xi, Y, \theta)],
\ee
where $ \QQ\opt=\QQ_0\otimes \QQ_1\opt \otimes \ldots \otimes \QQ_d\opt$,
and $\QQ_j\opt=(1-\delta)^{-1} \times \textrm{Bernoulli}(1-\delta)$ is a scaled Bernoulli distribution for any $j=1,\ldots,d$, i.e., under $\QQ_j\opt$
\begin{equation} \label{eqn:droputTheorem}
    \xi_j = \begin{cases}
    0 & \text{ with probability }  \delta, \\ (1-\delta)^{-1} & \text{ with probability }  (1-\delta).
    \end{cases}
\end{equation}
In addition, let $\theta\opt \in \Theta$ be a solution to  \eqref{eq:dro-refor}. Then $(\theta\opt,\QQ\opt)$ constitutes a Nash equilibrium of the two-player zero sum game defined by \eqref{eqn:loss_GLM} and \eqref{eqn:natures_choice} and $\QQ\opt$ is nature's least favorable distribution. 
\end{theorem}

\begin{proof}
See Appendix \ref{subsection:proofTheorem}.
\end{proof}

\noindent The first part of theorem characterizes the statistician's best response to an adversarial nature that is allowed to corrupt the covariates using a multiplicative errors-in-variables model. From the statistician's perspective, nature's worst-case perturbation of $\mathbb{Q}_0$ is given by $\QQ\opt$ in   \eqref{eqn:droputTheorem}. Under this \emph{worst-case} distribution, nature independently corrupts each of the entries of $X=(X_1, \ldots, X_{d})^\top$, by either dropping the $j$-th component (if $\xi_j=0$) or replacing it by $X_j/(1-\delta)$. Dropout training---which here refers to estimating the parameter $\theta$ after adding dropout noise to $X$---thus becomes the statistician's preferred way of estimating the parameter $\theta$ when facing an adversarial nature. This gives a decision-theoretic foundation for the use of dropout training. 

Note that in order to recover the objective function introduced in \eqref{eqn:Dropout_objective} (the sample average of the contaminated log-likelihood) it suffices to set the reference measure---$\mathbb{Q}_0$---as the empirical distribution $\widehat{\mathbb{P}}_n$ of $\{(x_i,y_i)\}_{i=1}^{n}$, which satisfies Assumption \ref{asn:reg}. Likewise, Theorem \ref{thm:glm} allows to interpret the probability limit $\beta\opt(\delta)$ of the dropout estimator derived in Proposition \ref{prop:limit} as a solution to a DRO problem. In particular
\begin{equation*}
    \beta\opt(\delta) =  \adjustlimits\argmin_{\vphantom{p}\beta} \sup_{\QQ \in\mathcal{U}(P^*,\delta) }\EE_{\QQ} \left[ -\ln f( Y | X \odot \xi, \beta, \phi)\right],
\end{equation*}
provided the data-generating distribution $P^*$ satisfies Assumption \ref{asn:reg}. In the specific case of the linear regression model, this means that we can interpret $\beta\opt(\delta)$ as giving us the population's best linear predictor for $y$ in terms of $x$, provided nature is allowed to perturb the distribution of covariates using a multiplicative error-in-variables model.

More generally, because dropout noise is nature's best response for every $\beta$, the dropout loss is an upper bound for the true loss:
\[ \EE_{\QQ\opt} \left[ -\ln f( Y | X \odot \xi, \beta\opt(\delta), \phi)\right] \geq \EE_{P^*} \left[ -\ln f( Y | X \odot \xi, \beta\opt(0), \phi)\right]. \]
We provide now some intuition about how dropout noise becomes nature's worst-case distribution. Algebra shows that, in light of Assumption \ref{asn:reg}, the expected loss under an arbitrary distribution $\QQ$ is finite and can be written as
\begin{align*}
    \EE_{\QQ}[\ell(X\odot\xi,Y,
    \theta)]&= -\mathbb{E}_{\mathbb{Q}_0} \left[  \ln h(Y,\phi) \right] \\
    & +  \mathbb{E}_{\mathbb{Q}_0}  \left[ \EE_{\QQ_1 \otimes \ldots \otimes \QQ_d}[(\Psi((\beta \odot X)^\top \xi )-Y((\beta \odot X)^\top \xi) )/a(\phi)] \right],
\end{align*}
where the first expectation is taken with respect to the reference distribution, and the second one with respect to $\xi$. For fixed values of $(X,Y,\theta)$ we can define 
\[ A_{(X,Y,\theta)} ( (\beta \odot X)^\top \xi ) \equiv (\Psi((\beta \odot X)^\top \xi )-Y((\beta \odot X)^\top \xi) )/a(\phi). \] 
Because $\Psi(\cdot)$ is a convex function defined on all of the real line, the function $A_{(X,Y,\theta)}(\cdot)$ inherits these properties. We show in the appendix that for these type of functions
\begin{equation} \label{eqn:key}
\sup\left\{ \EE_{\QQ_1 \otimes \ldots \otimes \QQ_d}[A_{(X,Y,\theta)} ( (\beta \odot X)^\top \xi )]: \QQ_j \in \mc Q_j(\delta)\right\} = \mathbb{E}_{\QQ_1\opt \otimes \ldots \otimes \QQ_d\opt}[A_{(X,Y,\theta)} ( (\beta \odot X)^\top \xi )],
\end{equation} 
for any $\theta$, and this establishes the equivalence between \eqref{eq:dro} and \eqref{eq:dro-refor}. The proof of the equality above  exploits convexity. In fact, to derive our result we first characterize the worst-case distribution for the expectation of a real-valued convex function (Lemma~\ref{lemma:convex}) and then we generalize this result to functions that depend on $\xi$ only through linear combinations, as $A_{(X,Y,\theta)}(\cdot)$ (Proposition~\ref{prop:exp}). 

 How about the Nash Equilibrium of the two-player zero sum game defined by \eqref{eqn:loss_GLM} and \eqref{eqn:natures_choice}? The equality in \eqref{eqn:key} clearly shows that $\QQ\opt$ is nature's best response for any $\theta \in \Theta$. If there is a vector $\theta\opt$ that solves the dropout training problem in \eqref{eq:dro-refor}, then this vector is the statistician best's response to nature's choice of $\QQ\opt$. Consequently, $(\theta\opt,\QQ\opt)$ is a Nash equilibrium. 

Finally, we discuss the extent to which $\QQ\opt$ can be referred to as nature's least favorable distribution, which has been defined as nature's solution to the maximin problem. It is well known that the maximin value of a game is always smaller than its minimax value:\footnote{This follows from the fact that for any $\QQ \in \mathcal{U}(\QQ_0,\delta):$
\[ \inf_{\theta \in \Theta} \EE_{\QQ} [\ell(X\odot\xi,Y,
    \theta) ] \leq \EE_{\QQ} [\ell(X\odot\xi,Y,
    \theta) ] \leq \sup_{\QQ \in \mathcal{U}(\QQ_0,\delta)} \EE_{\QQ} [\ell(X\odot\xi,Y,
    \theta) ].\] See also the discussion of the minimax theorem in \cite{Ferguson67} p. 81.
    } 
\[ \adjustlimits\sup_{\QQ \in \mathcal{U}(\QQ_0,\delta)} \inf_{\theta \in \Theta} \EE_{\QQ} [\ell(X\odot\xi,Y,
    \theta) ] \leq \adjustlimits\inf_{\theta \in \Theta} \sup_{\QQ \in \mathcal{U}(\QQ_0,\delta)} \EE_{\QQ} [\ell(X\odot\xi,Y,
    \theta) ].  \]


We have shown that the right-hand side of the display above equals \eqref{eq:dro-refor}. Therefore, if there is a $\theta\opt \in \Theta$ that solves such program, then $\QQ\opt$ achieves  the upper bound to the maximin value of the game. This makes dropout noise nature's least favorable distribution. 

Now that we have established that dropout training gives the minimax solution of the DRO game, we discuss the implications of this result regarding the out-of-sample performance of dropout training. Suppose $\QQ_0$ is the empirical measure $\wh \PP_n$ supported on $n$ training samples $\{(x_i, y_i)\}_{i=1}^n$. The \emph{in-sample} loss of dropout training is
\begin{equation} \label{eqn:in-sample-loss}
\frac{1}{n} \sum_{i=1}^{n} \EE_{\QQ\opt}[ \ell(X \odot \xi, Y, \theta\opt)|X=x_i,Y=y_i ].   
\end{equation}
A typical concern with estimation procedures is whether their performance in a specific sample translates to good performance out of sample. In our context, the out-of-sample performance of dropout training can be thought of as the expected loss that would arise for some other data distribution $\tilde{\QQ}_0$ over $(X,Y)$ at the parameter estimated via dropout training: 
\[
    \EE_{\tilde{\QQ}_0} [\ell(X,Y,\theta\opt)].  
\]
The minimaxity of dropout training shows that for any distribution $\tilde{\QQ}_0$ over $(X,Y)$ that can be obtained from $\widehat{\mathbb{P}}_n$ by perturbing covariates with mean-one independent multiplicative error $\xi_j \in [0,(1-\delta)^{-1}]$ we have
\[
    \EE_{\tilde{\QQ}_0} [\ell(X,Y,\theta\opt)] \leq \frac{1}{n} \sum_{i=1}^{n} \EE_{\QQ\opt}[ \ell(X \odot \xi, Y, \theta\opt) |X=x_i,Y=y_i].
\]
This means that the out-of-sample loss will be upper-bounded by the in-sample loss. Thus, our results give a concrete result about the class of distributions for which dropout training estimation ``generalizes'' well. 

Finally, we note that Theorem~\ref{thm:glm} was stated for a scalar $\delta$ that is homogeneous across the multiplicative noise $\xi_j$. To model non-identical dropout noise, we can substitute the sets in~\eqref{eqn:Qj} and~\eqref{eqn:natures_choice} by $\mc Q_j(\delta_j)$ for a collection of parameters $(\delta_1, \ldots, \delta_d) \in [0, 1)^d$. In this case, the results of Theorem~\ref{thm:glm} hold with $\QQ_j\opt=(1-\delta_j)^{-1} \times \textrm{Bernoulli}(1-\delta_j)$ for $j = 1, \ldots, d$.

%% file: 9New_Section_choicedelta.tex
Theorem \ref{thm:glm} in the previous section showed that dropout training is distributionally robust optimal and that  nature's least favorable distribution is  dropout noise with probability $\delta \in [0, 1)$. This section suggests a strategy to pick this parameter. Broadly speaking, our approach relies on a simple idea: we study how often the in-sample loss obtained from dropout training exceeds the population loss. If overfitting is successfully mitigated, this probability ought to be large. We show that by appropriately tuning the parameter $\delta$ of the dropout noise, it is possible to control the probability of such event as the sample size grows large.  

\subsection{Additional Notation}

Throughout this section, we use  $\widehat{\phi}$ to denote an arbitrary $\sqrt{n}$-consistent, asymptotically normal estimator for the scale parameter $\phi$. Such estimator can be obtained, for example, by using $\widehat{\phi}_{\mathrm{ML}}$ in~\eqref{eqn:MLE}. We use $\phi^\star$ to denote the true, unknown scale parameter. Just as before, we let $\widehat{\beta}(\delta)$ denote the dropout estimator of the true $\beta^\star$ under dropout probability $\delta$.

The in-sample loss of dropout training---given a dropout probability of $\delta$---evaluated at parameters $\beta$ and $\phi$ is given by
\begin{equation} \label{eqn:in_sample_loss}
\mathcal{L}_n(\beta, \phi, \delta) 
 \equiv \frac{1}{n} \sum_{i=1}^{n} \mathbb{E}_{\delta} \left[ -\ln f( y_i | x_i \odot \xi_i , \beta, \phi) \right].
\end{equation}
The goal of this section is to understand how likely is that the in-sample loss in (\ref{eqn:in_sample_loss})---when evaluated at the dropout estimator $\widehat{\beta}(\delta)$ and some estimator $\widehat{\phi}$---exceeds the true population loss. Thus, if we define the population loss as 
\begin{equation} \label{eqn:pop_loss}
\mathcal{L}(\beta^\star,\phi^\star) \equiv \mathbb{E}_{P^\star} \left[ -\ln f( Y | X, \beta^\star, \phi^\star) \right],
\end{equation}
we are interested in understanding how often 
\begin{equation} \label{eqn:event}
\mathcal{L}_{n} (\widehat{\beta}(\delta_n), \widehat{\phi}, \delta_n) \geq \mathcal{L}(\beta^\star,\phi^\star),
\end{equation}
where the sequence $\delta_n$ is allowed to change with the sample size.

\subsection{Additional Assumptions}
It is well-known that under mild regularity conditions on the true data generating process---and without the need of dropout training---the usual in-sample loss evaluated at the Maximum Likelihood estimators, which we can denote as  $\mathcal{L}_n(\widehat{\beta}_{\textrm{ML}}, \widehat{\phi}_{\textrm{ML}}, 0)$,  provides a consistent estimator for the population loss. This remains true if the Maximum Likelihood estimator for $\phi$ is replaced by another $\sqrt{n}$-consistent estimator $\widehat{\phi}$. One sufficient condition that guarantees such behavior is the following high-level assumption:
\begin{assumption} \label{assump:CLT} The following central limit theorem result holds for some $\sigma^2$
\[ \sqrt{n} \left( \mathcal{L}_n(\beta^\star, \widehat{\phi}, 0)  - \mathcal{L}(\beta^\star,\phi^\star)\right) \overset{d}{\rightarrow} \mathcal{N}(0, \sigma^2).  \] 
\end{assumption}
\noindent In particular, the identity 
\begin{align*}
\sqrt{n} \left( \mathcal{L}_n(\widehat{\beta}_{\textrm{ML}}, \widehat{\phi}, 0) -  \mathcal{L}(\beta^\star,\phi^\star) \right) = & \sqrt{n} \left( \mathcal{L}_n(\widehat{\beta}_{\textrm{ML}}, \widehat{\phi}, 0) -  \mathcal{L}_n(\beta^\star, \widehat{\phi}, 0) \right) \\ 
& + \sqrt{n} \left( \mathcal{L}_n(\beta^\star, \widehat{\phi}, 0)  -  \mathcal{L}(\beta^\star,\phi^\star)\right),
\end{align*}
and Assumptions \ref{assump:bounded}, \ref{assump:moments}, \ref{assump:CLT} imply
\begin{equation} \label{eqn:CLT_ML}
\sqrt{n} \left( \mathcal{L}_n(\widehat{\beta}_{\textrm{ML}}, \widehat{\phi}, 0) -  \mathcal{L}(\beta^\star,\phi^\star) \right) \overset{d}{\rightarrow} \mathcal{N}(0, \sigma^2).
\end{equation}
\noindent We view the result in (\ref{eqn:CLT_ML}) concerning the in-sample loss of Maximum Likelihood estimators, which is quite standard, as unsatisfactory. Our main complaint is that the probability of the event
\[  \mathcal{L}_n(\widehat{\beta}_{\textrm{ML}}, \widehat{\phi}, 0)  \leq \mathcal{L}(\beta^\star, \phi^\star) \]
approaches 1/2 as the sample size grows large. Our interpretation is that the in-sample loss at the Maximum Likelihood estimator is deceivingly small, as the true population loss will be above it 50\% of the time if the sample size is large enough. We argue that this probability can be made smaller by appropriately tuning dropout noise. 

\subsection{In-sample loss of dropout training}
Theorem \ref{thm:glm} showed that dropout noise is nature's choice to inflict the highest loss for the statistician at any parameter values. Therefore,  invoking Theorem \ref{thm:glm} using the empirical distribution as the reference distribution, the random variable 
\begin{equation} \label{eqn:mu_prop}
\mu_n(\beta, \phi, \delta) \equiv \mathcal{L}_n(\beta, \phi,\delta) - \mathcal{L}_n(\beta, \phi,0) 
\end{equation}
is nonnegative for any $\delta \in [0,1)$. Since Proposition \ref{prop:limit} has shown that $\widehat{\beta}(\delta_n) \overset{p}{\rightarrow} \beta^\star$ for any sequence $\delta_n \rightarrow 0$, intuition suggests that the in-sample loss of dropout training, $\mathcal{L}_{n} (\widehat{\beta}(\delta_n), \widehat{\phi}, \delta_n)$, may provide a consistent estimate of the population loss that does not underestimate this limit frequently.

Our result is the following proposition: 

\begin{proposition} \label{prop:loss}
Suppose that Assumptions \ref{assump:bounded}, \ref{assump:moments}, \ref{assump:CLT} hold. Then for any sequence  $\delta_n=c/\sqrt{n}$,
\[ \sqrt{n} \left(   \mathcal{L}_{n} (\widehat{\beta}(\delta_n), \widehat{\phi}, \delta_n) - \mathcal{L}(\beta^\star,\phi^\star) \right) \overset{d}{\rightarrow} \mathcal{N} ( \mu_{\infty}(\beta^\star,\phi^\star,c), \sigma^2),  \]
\noindent where $\mu_{\infty}(\beta^\star,\phi^\star,c) \geq 0$ is the probability limit of 
\[ \sqrt{n} \mu_n \left( \beta^\star ,\widehat{\phi}, \delta_n  \right),\]
and $\mu_n (\cdot)$ is defined as in (\ref{eqn:mu_prop}).
\end{proposition}

\begin{proof}
See Appendix \ref{subsection:loss}.
\end{proof}

The main message of Proposition \ref{prop:loss} is that the probability of the event (\ref{eqn:event}) can be approximated, as the sample size goes large, by the probability---under a normal random variable with positive mean---of the positive half of the real line. 
Some elementary algebra can be used to illustrate the main argument behind the proof. Note that
\begin{align*}
  \sqrt{n} \left(   \mathcal{L}_{n} (\widehat{\beta}(\delta_n), \widehat{\phi}, \delta_n) - \mathcal{L}(\beta^\star,\phi^\star) \right) &=   \sqrt{n} \left(   \mathcal{L}_{n} (\widehat{\beta}(\delta_n), \widehat{\phi}, \delta_n) - \mathcal{L}_n(\beta^\star,\widehat{\phi}, \delta_n) \right) \\
  &+  \sqrt{n} \mu_n(\beta^\star, \widehat{\phi}, \delta_n) \\
   &+ \sqrt{n} \left(   \mathcal{L}_n(\beta^\star,\widehat{\phi}, 0) - \mathcal{L}(\beta^\star,\widehat{\phi}) \right).
\end{align*}
We start by showing that the first term converges in probability to zero. To do this we show that  
\[\sqrt{n}(\widehat{\beta} (c/\sqrt{n}) - \beta^\star)\] 
is asymptotically normal and that the derivative of $\mathcal{L}_n(\cdot)$ with respect to $\beta$ (evaluated at $\beta^\star$) converges in probability to zero.  

The key step in the proof shows that the second term has a finite probability limit. In fact, we can characterize this limit explicitly and show that
\[ \sqrt{n}\mu_n(\beta^\star,\widehat{\phi},\delta_n) \overset{p}{\rightarrow} c \cdot \mu,\] 
where
\[ \mu \equiv \left( \left( \sum_{\xi\in \mathcal{A}}\mathbb{E}_{P^\star}[\Psi((X\odot\xi)^{\top} \beta^\star)  ] \right) -d\mathbb{E}_{P^\star}[\Psi(X ^\top \beta^\star )]+\mathbb{E}_{P^\star}[Y X^\top  ]\beta^\star \right) \Big/ a(\phi^\star),
\]
and $\mathcal{A}$ is the collection of all vectors in $\{0,1\}^{d}$ for which there is only one zero. In Section \ref{sec:numerics} we provide an expression for this term in the linear regression model.

It is important to mention that the DRO interpretation of dropout training can be leveraged to select the dropout parameter $\delta$. For example, a possible approach consists in choosing $\delta$ so that true data generating process belongs to nature's choice set with some prespecified probability. This approach, which is often advocated in the literature in machine learning and robustness \citep{ref:hansen2008robustness}, often leads to a very pessimistic selection of $\delta$ simply because this criterion is not informed at all by the loss function defining the decision problem. Further, in our problem, it is not possible to apply this approach given that the set of multiplicative perturbations of the empirical distribution will, in general, not cover the true data generating process.

Another approach involves using  generalization bounds leading to finite sample guarantees; see, for instance a summary of this discussion in Section 6.2 of \cite{RM2019}. This method, while appealing, often requires either distributions with compact support or strong control on the tails of the underlying distributions. Also, often, the bounds depend on constants that may be too pessimistic or difficult to compute. 

Finally, there is a recent method introduced in \cite{BKM2016} for the case in which nature's choice set is defined in terms of the Wasserstein's distance around the empirical distribution. The idea therein is that---for a fixed $\delta$---every distribution that belongs to  nature's choice set corresponds to an optimal parameter choice for the statistician. Thus, one can collect each and every of the statistician's optimal choices associated to each distribution in nature's uncertainty set, and treat the resulting region as a confidence set for the true parameter. This confidence set grows bigger (in the sense of nested confidence regions) as $\delta$ increases. The goal is then to minimize $\delta$ subject to a desired level of coverage in the underlying parameter to estimate. This leads to a data-driven choice of $\delta$ that is explicitly linked to the statistician's decision problem. However, this approach is not feasible in our problem because, once again, regardless of the value of $\delta$, the parameter choices for each of the multiplicative perturbations of the empirical distribution will, in general, fail to cover the true parameter.   

Hence, we advocate the strategy of choosing the parameter $\delta$ to control how often the in-sample loss obtained from dropout training exceeds the population loss. The proof of Proposition \ref{prop:loss} shows that $\mu_\infty(\beta^\star,\phi^\star,c)$ is of the form $c \cdot \mu$, where $\mu$ depends on $(\beta^\star,\phi^\star)$. Consequently, as long as $\mu>0$, it is straightforward to pick $c$ to guarantee a pre-specified ``coverage'' of the population loss: for any $\alpha\in(0,1)$, if we pick $c$ to be
\[ z_{1-\alpha}  \cdot \sigma / \mu, \]
 where $z_{1-\alpha}$ is the 1-$\alpha$ quantile of a standard normal, then the probability of the event \eqref{eqn:event} asymptotically approaches $1-\alpha$. Thus, overfitting can be successfully mitigated.

%% file: 5Algorithm.tex
The goal of this section is to suggest an algorithm for solving the dropout training problem
\[ \inf_{\theta \in \Theta} \EE_{\QQ\opt}  [\ell(X\odot\xi,Y,
    \theta)  ], \]
where $\QQ\opt =\wh \PP_n \otimes \QQ_1\opt \otimes \ldots \otimes \QQ_d\opt$ and $\QQ\opt_{j}$, $j=1,\ldots, d$ is the dropout noise distribution defined in \eqref{eqn:droputTheorem}. Notice that we here consider the specific case in which $\QQ_0$ is set to the empirical measure $\wh \PP_n$ supported on $n$ training samples $\{(x_i, y_i)\}_{i=1}^n$.  We will use $\theta_{n}\opt$ to denote the solution of the dropout training problem above. It will sometimes be convenient to rewrite this dropout training problem as
\begin{equation} \label{eqn:dropout_conditioning}
    \min_{\theta} \frac{1}{n} \sum_{i=1}^{n} \EE_{\QQ\opt} [ \ell (X\odot \xi,Y,\theta) \: | \: X= x_i, Y= y_i ], 
\end{equation}
which coincides with expression \eqref{eqn:Dropout_objective}. Conditioning on the values of $(x_i, y_i)$ makes it clear that the expectation is computed over the $d$-dimensional vector $\xi$. We now briefly describe three common approaches to implement dropout training and we discuss some of its limitations.

\subsection{Naive Dropout Training} Because $\QQ_j\opt$ places mass on only two points, namely $0$ and $(1-\delta)^{-1}$, the support of the joint distribution $\QQ_1\opt \otimes \ldots \otimes \QQ_d\opt$ has cardinality $2^d$. Thus, a naive approach to solve the dropout training problem~\eqref{eqn:dropout_conditioning} is to expand the objective function as a sum with $n\cdot2^d$ terms, then to apply a tailored gradient descent algorithm to the resulting optimization problem. Unfortunately, this approach is computationally demanding because the number of individual terms in the objective function grows exponentially with the dimension $d$ of the features. 

\subsection{Dropout Training via Stochastic Gradient Descent}\label{sec:vanillasgd} Another method to solve the dropout training problem in~\eqref{eq:dro-refor} is by stochastic gradient descent (henceforth, SGD). This gives us the commonly used dropout training algorithm. For the sake of comparison, we provide concrete details about this algorithm below.

Given a current estimate $\wh\theta$, we compute an unbiased estimate of the gradient to the objective function of~\eqref{eq:dro-refor}, and move in the direction of the negative gradient with a suitable step size. Since $\QQ\opt$ is discrete, the expectation under $\QQ\opt$ can be written as a finite sum and by differentiating under the expectation, we have 
\begin{equation}\label{eq:truegrad}
 \nabla_{\theta} \EE_{\QQ\opt}[\ell(X\odot\xi, Y, \wh \theta )]=\EE_{\QQ\opt}\left[\nabla_{\theta}\ell(X\odot\xi, Y, \wh \theta )\right].
\end{equation}
 The standard SGD algorithm uses a naive Monte Carlo estimator as an estimate of the gradient~\eqref{eq:truegrad}, that is, at iterate $k \in \mathbb{N}$ with incumbent solution $\wh \theta^k$,
 \[
  \nabla_\theta \EE_{\QQ\opt}[\ell(X\odot\xi, Y, \wh \theta^k )] \approx \nabla_{\theta} \ell(x_k \odot \xi_k, y_k, \wh \theta^k),
\]
where $(x_k,y_k,\xi_k)$ is an independent draw from $\QQ\opt$.
 
One drawback of using SGD to solve~\eqref{eq:dro-refor} is that it is not easily parallelizable, and thus its implementation can be quite slow. Moreover, under strong convexity assumption of the loss function $\ell$, SGD only exhibits linear convergence rate~\cite[Section 2.1]{Nemirovski2009}. By contrast, the gradient descent (GD) enjoys exponential convergence rate~\cite[Section 9.3.1]{Boyd2004}.

\subsection{Naive Monte Carlo Approximation for Dropout Training} Consider solving the dropout training problem in \eqref{eqn:dropout_conditioning} using a naive Monte Carlo approximation. Instead of using $2^d$ terms to compute 
\[  \EE_{\QQ\opt} [ \ell (X\odot \xi,Y,\theta) \: | \: X= x_i, Y= y_i ], \] 
we approximate this expectation by taking a large number of $K$ i.i.d.~draws $\{\xi_i^k\}_{k=1}^{K}$,  ${\xi^k_i \in \mathbb{R}^d}$, according to the distribution $\QQ_1^* \otimes \ldots \otimes \QQ_d^*$. When $d$ is large this approximation is computationally cheaper than the naive dropout training procedure described above, provided that $K \ll 2^d$. 

Thus, the naive Monte Carlo approximation of the dropout training problem is
\begin{equation}\label{eq:sampopt}
    \min_{\theta\in \Theta}~\frac{1}{n} \sum_{i=1}^{n} \left[ \frac{1}{K}\sum_{k=1}^K \ell(x_i\odot\xi^k_i,y_i,\theta) \right],
\end{equation}
where the random vectors $\xi_{i}^k$ are sampled independently---over both $k$ and $i$---using the distribution $\QQ_1\opt \otimes \ldots \otimes \QQ_d\opt$. 

Relative to the solution of the dropout training problem---which we denoted as $\theta_{n}\opt$---the minimizer of~\eqref{eq:sampopt} is consistent and  asymptotically normal as $K \rightarrow \infty$. This follows by standard arguments; for example, those in~\citet[Section 5.1]{Shapiro:2014}. There are, however, two problems that arise when using \eqref{eq:sampopt} as a surrogate for the dropout training problem. First, the solution to \eqref{eq:sampopt} is a biased estimator for $\theta\opt_n$. This means that if we average the solution of \eqref{eq:sampopt} over the $K \cdot n$ different values of $\xi^k_{i}$, the average solution need not equal $\theta\opt_n$. Second, implementing \eqref{eq:sampopt} requires a choice of $K$ and, to the best of our knowledge, there is no off-the-shelf  procedure for picking this number.   

\subsection{Unbiased Multi-level Monte Carlo Approximation for Dropout Training}\label{sec:saaunbiased} 

To address these two issues, we apply the recent techniques suggested in~\cite{ref:blanchet2019unbiased} that we refer to as \emph{Unbiased Multi-level Monte Carlo Approximations}. Multi-level Monte Carlo methods \citep{Giles:2008,Giles:2015} refer to a set of techniques for approximating the expectation of random variables. The adjective ``multi-level'' emphasizes the fact that random samples of different \emph{levels} of accuracy are used in the approximation. Before presenting the detailed algorithm, we provide a heuristic description. To this end, let $\widehat{\theta}\opt_{n}(K)$ denote the \emph{level} $K$ solution of the problem in \eqref{eq:sampopt}; that is, the solution based on $K$ draws. Define the random variable 
\[ \Delta_{K} \equiv \widehat{\theta}\opt_{n}(K) - \widehat{\theta}\opt_{n}(K-1). \]
and, for simplicity, assume $\widehat{\theta}\opt_n (0)$ is defined to equal a vector of zeros. Under suitable regularity conditions, there holds
\[ \sum_{K=1}^{\infty} \EE [ \Delta_{K} ]  = \lim_{K \rightarrow \infty} \EE [ \widehat{\theta}\opt_{n}(K) ] =   \theta\opt_n.  \]
Consider now picking $K^*$ at random from some discrete distribution supported on the natural numbers. Let $p(\cdot)$ denote the probability mass function of such distribution and consider a Monte Carlo approximation scheme in which---after drawing $K^*$---we sample $K^* \cdot n$ different random vectors $\xi^{k}_{i} \in \R^d$ according to $\QQ_1\opt \otimes \ldots \otimes \QQ_d\opt$. The estimator 
\[ Z(K^*) \equiv \frac{\Delta_{K^*}}{p(K^*)} \]
has two sources of randomness. Firstly, the random choice of $K^*$ and, secondly, the random draws $\xi^{k}_{i}$. Averaging over both yields
\[ \EE[Z(K^*)] = \sum_{K=1}^{\infty} \EE[ Z(K^*) | K^* = K   ] \cdot p(K) = \sum_{K=1}^{\infty} ( \EE [ \Delta_{K} ] / p(K))  \cdot p(K) = \theta\opt_n. \]
Thus, by taking into account the randomness in the selection of $K$, we have managed to provide a rule for deciding the number of draws (specifically, our recommendation is to pick $K^*$ at random) and at the same time we have removed the bias of naive Monte Carlo approximations. 


One possible concern with our suggested implementation is that the expected computational cost of $Z(K^*)$ could be infinitely large. Fortunately, this issue can be easily resolved by an appropriate choice of the distribution $p(\cdot)$. To see this, define the computational cost simply as the number of random draws that are required to  obtain $Z(K^*)$. In the construction we have described above, we need $K^* \cdot n$ draws for the construction of the estimator. Thus, the average cost is 
\[ \mathbb{E}[ K^* \cdot n ] = n \sum_{K=1}^{\infty} K \cdot p(K)   \]
which, under mild integrability conditions on $p(\cdot)$, will be finite.\footnote{For example, if $p(\cdot)$ is selected as a geometric distribution with parameter $r$, the expected computational cost will be $n(1-r)/r$. }

We now present the algorithm that will be used to solve the dropout training problem. To ensure that the estimator $Z(K^*)$ has a finite variance, instead of defining $\Delta_K$ as the difference between the level $K$ and $K-1$ solutions to problem \eqref{eq:sampopt} in the above heuristic arguments, we use solutions to problem \eqref{eq:sampopt} with a sample of size $2^{K+1}$ and with its \textit{odd} and \textit{even} sub-samples of size $2^K$. 

{\scshape Algorithm for the Unbiased Multilevel Monte Carlo:} We present a parallelized version of it using $L$ processors, but the suggested algorithm works even when $L=1$. Parallel computing reduces the variance of the estimator, and our suggestion is to use as many processors as available in one run.  

Fix an integer $m_0 \in \mathbb{N}$ such that $2^{m_0+1} \ll 2^d$. For each processor $l=1, \ldots, L$ we consider the following steps. 

\begin{enumerate}
\item [i)] Take a random (integer) draw, $m_l^*$, from a geometric distribution with parameter $r>1/2$.\footnote{To see why we require that $r>1/2$, notice if the computational cost of evaluating $Z(K^*)$ (as in the heuristic description above) increases exponentially in $K$ and takes the form $C \cdot 2^{K}$, the expected computational cost will be 
\[ \sum_{K=1}^{\infty} C r (2(1-r))^K = C r (1/2(1-r)), \]
provided $2(1-r)<1$, or equivalently, $r>1/2$. As we show in the proof of Theorem \ref{thm:unbias}, constraining the variance requires then imposing $r<3/4$. Ultimately, optimizing the product of computational cost and variance leads to the optimal selection $r=1-2^{-3/2}$.} 

\item [ii)] Given $m^*_l$, take $2^{K^*_{l}+1}$ i.i.d.~draws from  the $d$-dimensional vector $\xi_i \sim \QQ_1\opt \otimes \ldots \otimes \QQ_d\opt$, where 
\[
    K_{l}^* \equiv m_0 + m_l^*.
\]
Repeat this step independently for each $i=1, \ldots, n$. 

\item [iii)] Solve problem \eqref{eq:sampopt} using the first $2^{m_0}$ i.i.d.~draws of $\xi_i$ for each $i$. Let $\theta_{l, m_0}$ denote a minimizer. 

\item [iv)] Denote by $\widehat{\theta}\opt_{n} ( 2^{K^*_{l}+1} )$, $\widehat{\theta}^{O}_{{n}}( 2^{K^*_{l}})$, and $\widehat{\theta}^{E}_{n}( 2^{K^*_{l}} )$ any solution to the following optimization problems (all of which are based on sample average approximations as \eqref{eq:sampopt}):
\begin{align*}
  \widehat\theta^\star_{n} ( 2^{K^*_{l}+1} ) & \in \arg\min_{\theta\in \Theta} \frac{1}{n} \sum_{i=1}^{n} \left( \frac{1}{2^{K_l^*+1}}\sum_{k=1}^{2^{K_l^*+1}}\ell(x_i\odot\xi^k_i, y_i,\theta) \right) , \\
  \widehat{\theta}^{O}_{{n}}(2^{K^*_{l}}) & \in \arg\min_{\theta\in \Theta} \frac{1}{n} \sum_{i=1}^n \left( \frac{1}{2^{K^*_l}}\sum_{k=1}^{2^{K_l^*}}\ell(x_i\odot\xi^{2k-1}_i,y_i,\theta) \right) , \\
  \widehat{\theta}^{E}_{n}( 2^{K^*_{l}} ) & \in \arg\min_{\theta\in \Theta}\frac{1}{n} \sum_{i=1}^{n}  \left( \frac{1}{2^{K_l^*}}\sum_{k=1}^{2^{K_l^*}}\ell(x_i\odot\xi^{2k}_i,y_{i},\theta) \right).
\end{align*}
Intuitively, $\widehat{\theta}_{n}^{O}$ and $\widehat{\theta}_{n}^{E}$ denote the solutions to problem \eqref{eq:sampopt} but using a sample of size $2^{K_l}$ with only  \textit{odd} and \textit{even} indices, respectively. 
\item [v)] Define
\[
   \bar\Delta_{K_l^*} \equiv  \widehat\theta^{\star}_{n} ( 2^{K^*_{l}+1} )-\frac{1}{2}( \widehat{\theta}^{O}_{n}( 2^{K^*_{l}} )+\widehat{\theta}^{E}_{n}( 2^{K^*_{l}} ))
\]
and let
\[
    Z (K^*_l) = \frac{\bar\Delta_{{K_l^*}}}{ r (1-r)^{K^*_l-m_0} } +\theta_{l,{m_0}}.
\]

\end{enumerate}

Our recommended estimator is 
\[
\frac{1}{L} \sum_{l=1}^{L} Z(K^*_l). 
\]

We now show that the suggested algorithm gives an estimator with desirable properties. We do so under the following regularity assumptions. 

\begin{assumption}\label{assump:com}
Suppose that the  parameter space $\Theta$ is compact. Suppose in addition that the optimal solution $\theta_n^\star$ to the dropout training problem in \eqref{eqn:dropout_conditioning} is (globally) unique. 
\end{assumption}

\begin{assumption}\label{assump:L2}
Let  $\widehat\theta^{\star}_n(K)$ denote the solution of the problem in \eqref{eq:sampopt} based on $K$ draws. Suppose that as $K \rightarrow \infty$,
\[
\EE[\|K^\half(\widehat\theta_n^\star(K)-\theta_n^\star)\|_2^4)=O(1),\]
where the expectation is taken over the i.i.d dropout noise distribution used to generate $\xi^k_{i}$.

\end{assumption}

\begin{assumption}\label{assump:smooth} Assume that for each $(X,Y,\xi)$, $\ell (X\odot \xi,Y,\cdot)$ is thrice continuously differentiable over $\Theta$ and that 
\[
\nabla_{\theta\theta}\EE_{\QQ^\star}[\ell(X\odot\xi,Y,\theta_n^\star)]
\]
is non-singular.

\end{assumption}

\begin{theorem}\label{thm:unbias}
Under Assumption~\ref{assump:com}, $\EE [Z(K^*_l)]=\theta_n^\star$. The number of random draws required to compute $Z(K^*_l)$ is $n \cdot 2^{K^*_l+1}$ and thus the  expected computational complexity for producing $Z(K^*_l)$ equals 
\[  \frac{n (2^{m_0+1}) r}{2r-1} < n (2^{m_0+1}) \ll n 2^d.   \]
Suppose, in addition, that  $\widehat\theta_n^\star(K)$ is almost surely in the interior of $\Theta$ for $K$ large enough. If Assumptions~\ref{assump:L2} and \ref{assump:smooth} hold, and $r<3/4$.  Then $\mathrm{Var}(Z(K^*_l))<\infty$.
\end{theorem}

\begin{proof}
See Appendix \ref{subsection:Proof_Bias}.
\end{proof}

Our suggested algorithm has finite  expected computational complexity that does not grow exponentially with the dimension $d$, thus every time we need to obtain $\widehat\theta\opt_n(2^{K^\star_l+1})$, we can do so by applying a gradient descent algorithm. Combined with parallelization, the Unbiased Multi-level Monte Carlo approach produces an unbiased estimator with a variance that can be made arbitrarily small if $L$ is large enough, provided that the regularity assumptions that give $\mathrm{Var}(Z(K^*_l))< \infty$ are satisfied.


%% file: 6Numerics.tex
We conduct numerical experiments in this section to compare our preferred implementation of dropout training to Stochastic Gradient Descent, as well as our recommended selection of $\delta$ to cross-validation. The benefits of our suggested Unbiased Multi-Level Monte Carlo algorithm are analyzed using a high-dimensional regression, whereas our selection of $\delta$ is analyzed using a low-dimensional regression model.

\subsection{Advantage of the Unbiased Multi-level Monte Carlo Estimator}
We present a simple numerical experiment to illustrate the advantage of using the Unbiased Multi-level Monte Carlo estimator suggested in Section~\ref{sec:saaunbiased}. We consider the linear regression problem with known variance and we focus on solving the dropout training problem with our recommended $\delta$ chosen according to Proposition~\ref{prop:loss}.

Our simulation setting considers a linear regression model with covariate vector having dimension $d=100$ and sample size $n=50$. We pick a known regression coefficient $\beta_0\in\mathbb{R}^d$ being a vector with all entries equal to 1. 
With fixed coefficients, we assume the covariate vector follows independent Gaussian, as well as for the regression noise. More specifically, we can get our $n=50$ observations $(x_i, y_i)$ via
\begin{itemize}
\item sampling $ x_i \sim\mathcal{N}(0,I_{d})$, $i=1,\ldots,n$,
\item sampling $ y_i\in\mathbb{R}$ conditional on $x_i$, where $y_i$ is given by the linear assumption and $\eps_i$ are i.i.d.~random noise following $\mathcal{N}(0,10^2)$, for $i=1,\ldots,n$. 
\end{itemize}
Our simulation setting considers first a high-dimension setting (relative low ratio between sample size per dimension $n/d=0.5$) with high noise to signal ratio (variability on residual noise is high compared to the variability on $x_i$). 

If we set $\QQ_0$ to be the empirical distribution of $\{(x_i,y_i)_{i=1}^{n}\}$, the dropout training problem in the linear regression model is
\[
\min_{\beta\in\mathbb{R}^d}~ \mathbb{E}_{\QQ\opt}\left[\left(\beta^\top (X\odot \xi)-Y\right)^2\right].
\]
Corollary~\ref{corol:lr} in Appendix~\ref{subsection:lr} shows that in the linear regression model the dropout training problem can be written as
\[ 
\min_{\beta\in\mathbb{R}^d}~\frac{1}{n}\left[(\mathbf{Y}-\mathbf{X}\beta)^\top(\mathbf{Y}-\mathbf{X}\beta)+\frac{\delta}{1-\delta}\beta^\top \mathbf{\Lambda}\beta\right],
\] 
where $\mathbf{Y} = [ y_1, y_2,\ldots,  y_n]^\top$, $\mathbf{X} = [ x_1,  x_2,\ldots, x_n]^\top$ and $\mathbf{\Lambda}$ is the diagonal matrix with its diagonal elements given by the diagonals of $\mathbf{X}^\top\mathbf{X}$. Moreover, there is a closed-form solution for the dropout training problem and it is given by the ridge regression formula:
\[
\beta^\star_n=\left(\mathbf{X}^\top\mathbf{X}+\frac{\delta}{1-\delta}\mathbf{\Lambda}\right)^{-1}\mathbf{X}^\top\mathbf{Y}.
\]

We choose the dropout probability $\delta$ following Proposition~\ref{prop:loss}. More specifically, Proposition~\ref{prop:loss} suggests the choice $\delta = c/\sqrt{n}$ where $c= z_{1-\alpha}\cdot\sigma/\mu$. For linear regression with known variance, it is straightforward to compute
\[
\mu = \frac{1}{2\phi^\star}\sum_{j=1}^d\mathbb{E}_{P^\star}[X_j^2](\beta_j^\star)^2,
\]
and
\[
\sigma^2 = \mathbb{V}\mathrm{ar}_{P^\star}\left[\frac{1}{2}\log(2\pi\phi^\star)+\frac{(Y-(\beta^\star)^\top X)^2}{2\phi^\star} \right].
\]
Choosing $\alpha=0.1$ and note that $\beta^\star=\beta_0, \phi^\star=10^2$, we have $\delta\approx 0.26$.

Since neither our suggested Multi-level Monte Carlo algorithm nor standard SGD (as defined in Section~\ref{sec:vanillasgd}) uses closed-form formulae for their implementation, we analyze the extent to which these procedures can approximate the parameter $\beta^\star_n$. We provide more details of the algorithms as follows. The two algorithms we compare are:
\begin{itemize}
\item Standard SGD algorithm with a learning rate $0.0001$, and initialization at the origin. Note that however we take batched SGD instead of single-sample SGD introduced in Section~\ref{sec:vanillasgd}.
\item Multi-level Monte Carlo algorithm with the geometric rate $r=0.6$ and the burn-in period $m_0 = 5$. Note that in each parallel running, we use gradient descent (GD) with $0.01$ learning rate and initialization at origin for steps iii) and iv) in Section~\ref{sec:saaunbiased}.

\end{itemize}
We run our simulation on a cluster with two Intel(R) Xeon(R) CPU E5-2640 v4 @ 2.40GHz processors (with 10 cores each), and a total memory of 128 GB. 
We fix 60 seconds as a ``wall-clock time'', so that we terminate the two algorithms after 60 seconds.\footnote{The parameters for the SGD algorithm are appropriately tuned to achieve good convergence within 60s (see Appendix~\ref{subsection:addnum} for the tuning procedure). However, we do not claim that this choice of parameters is optimal.}
We run $1000$ independent experiments. For each run, we calculate and report the average parameter estimation divergence to $\beta^\star_n$ and 1-standard deviation error bar for the divergence.
We consider difference number of parallelizations (i.e., $L$ in Section~\ref{sec:saaunbiased}) from 400 to 2400. We cap the run at $2400$ due to the saturation of divergence after $\sim 2000$ parallelizations. 

\begin{figure}[h]
    \centering
    \includegraphics[width=0.6\textwidth]{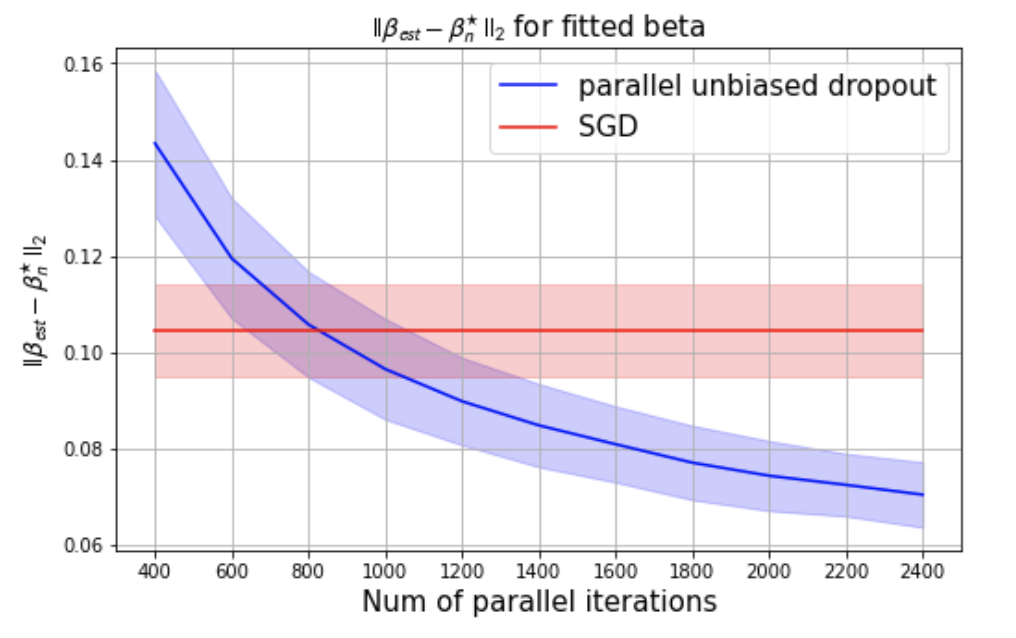}
    \caption{$l_2$ difference}
    \label{fig:1}
\end{figure}
\begin{figure}[h]
    \centering
    \includegraphics[width=0.6\textwidth]{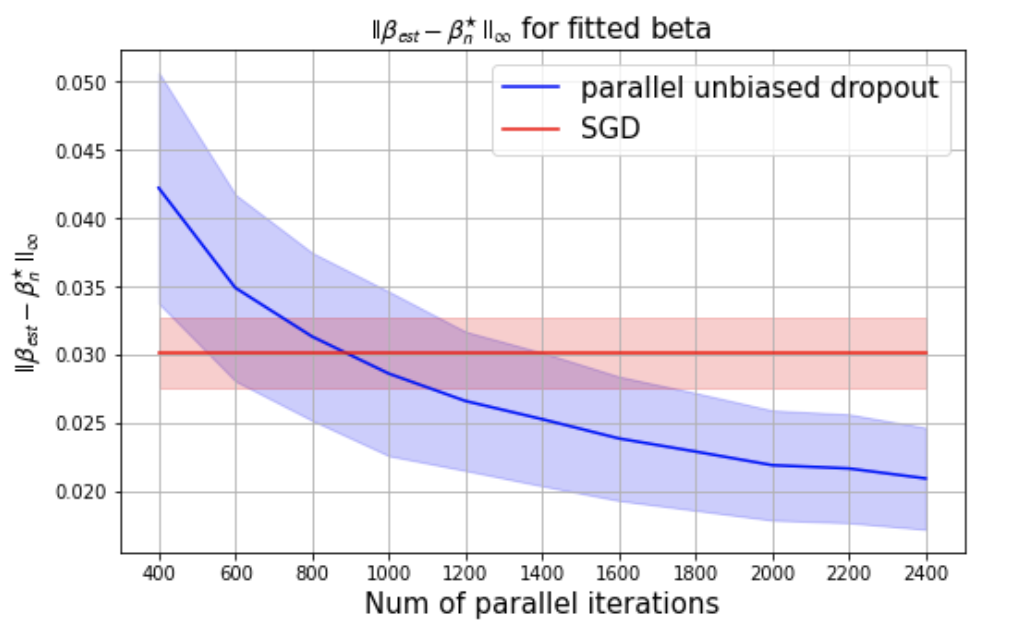}
    \caption{$l_\infty$ difference}
    \label{fig:2}
\end{figure}
\begin{figure}[h]
    \centering
    \includegraphics[width=0.6\textwidth]{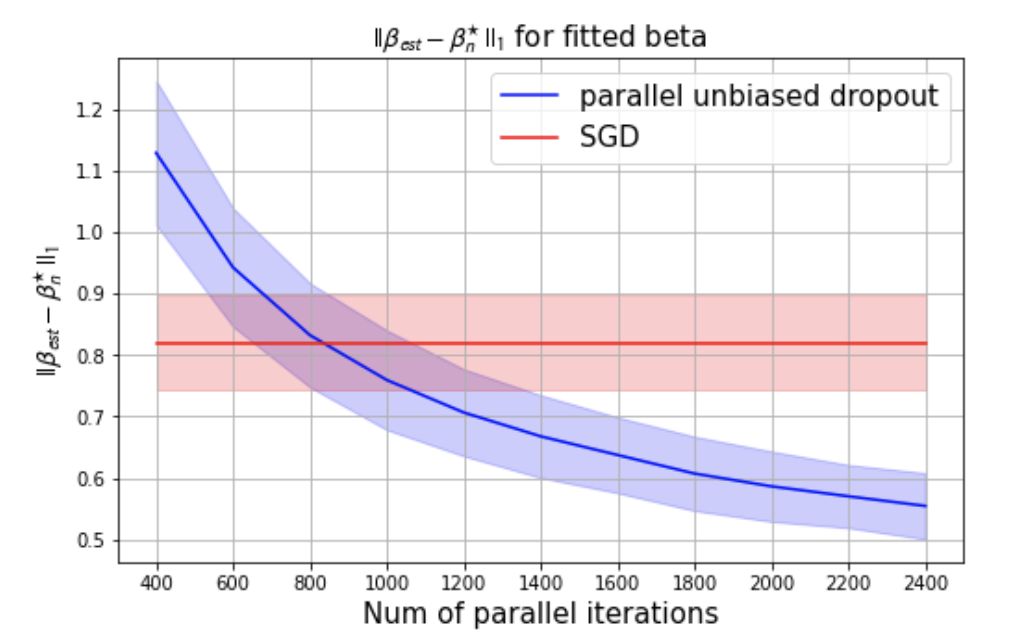}
    \caption{$l_1$ difference}
    \label{fig:3}
\end{figure}

Figure~\ref{fig:1} shows the $l_2$ divergence to the true $\beta^\star_n$ of the two algorithms for varying $L$, while Figure~\ref{fig:2} and Figure~\ref{fig:3} show $l_\infty$ and $l_1$ divergence, respectively. We observed that our unbiased estimator outperforms standard SGD algorithm once the number of parallel iterations reaches above some moderate threshold ($\sim 1000$ here). We provide supporting evidence in Appendix~\ref{subsection:addnum} to argue our choice of learning rate, initialization, and wall-clock time, where our proposed algorithm is robust to any reasonable choices.

\subsection{Coverage of the True Loss of Dropout Training}
We validate that our recommended selection of $\delta$ guarantees that the in-sample loss of dropout training is covering the true loss with arbitrary high probability as prescribed by Proposition~\ref{prop:loss}.

We use the same linear regression model with dimension $d=10$ and training samples $n \in \{10^3,10^4\}$.  We choose different quantiles of the normal as in Proposition~\ref{prop:loss}. We also include $10$-fold cross validation and ordinary least squares for comparison. See Table~\ref{tab:losscoverage}, where we estimate the frequency of coverage over $1000$ independent runnings. The main message is that our suggested choice of $\delta$ guarantees that the in-sample loss of dropout training exceeds the true, unknown, population loss with probability $1-\alpha$. Using standard OLS or choosing $\delta$ by cross-validation the in-sample loss is smaller than the population loss with probability close to 1/2, which implies that these methods are unsatisfactory in terms of frequency of coverage.

\begin{table}[h!]
\centering
\begin{tabular}{|c| c c c c c|} 
 \hline
   & $\alpha = 0.2$ & $\alpha=0.1$ & $\alpha=0.05$ & $10$-fold CV & plain OLS\\
 \hline
 $n=10^3$ & $0.77\pm0.01$ & $0.88\pm0.01$ & $0.94\pm0.01$ & $0.52\pm0.02$ & $0.40\pm0.01$ \\
 $n=10^4$ & $0.79\pm0.01$ & $0.90\pm0.01$ & $0.94\pm0.01$ & $0.49\pm0.02$ & $0.47\pm0.02$ \\
 \hline
\end{tabular}
\caption{Frequency of in-sample loss covering the true population loss. Our recommended selection of $\delta=c/\sqrt{n}$ with $c = z_{1-\alpha}\sigma/\mu$ has a theoretical $1-\alpha$ coverage probability. }
\label{tab:losscoverage}
\end{table}

%% file: 7Extensions.tex
In this section we discuss the extent to which the decision-theoretic support for dropout training carries over to Neural Networks. The main idea is that we use a GLM model where the natural parameter is no longer a linear function of the covariates, but instead a neural network.

\subsection{One-hidden-layer Feed-Forward Neural Networks} Suppose the scalar response variable $Y$ is generated by the conditional density
\begin{equation} \label{eqn:NN}
    f(Y|X, \theta,\phi) \equiv h(Y,\phi)\exp\left(\left(Y \Omega_{\theta}(X) )-\Psi(\Omega_{\theta}(X))\right)/a(\phi)\right),
\end{equation}
where $\Omega_{\theta}(X)$ is a neural network with parameters $\theta$ and $X \in \mathbb{R}^d$. This is a simple extension of the regression model that has been used recently to study deep neural networks; see \cite{schmidt2017nonparametric} in which the conditional density is Gaussian. 

In this section, we will assume that  $\Omega_{\theta}(X)$ is a neural network with a \emph{single hidden layer}, a  \emph{differentiable activation (squashing) function}, and \emph{linear ouput function}. A function $h:\mathbb{R} \rightarrow [0,1]$ is a squashing function if it is non-decreasing and if 
\[ \lim_{r \rightarrow \infty} h(r) = 1, \quad \lim_{r \rightarrow -\infty} h(r) = 0. \]
See Definition 2.3 in \cite{Hornik:1989}.

Although these types of networks---which will be formally described below---are  restrictive compared to the modern deep learning architectures, they can approximate any Borel measurable function from a finite-dimensional space to another, provided the hidden units in the hidden layer are large; see \cite{Hornik:1989}.

Consider a neural network with $K$ units in the hidden layer, each using input weights $w_{k} \in \mathbb{R}^{d}$, $k=1,\ldots, K$. Denote the activation function in the hidden layer as $h(\cdot)$. Assume the output function is linear with vector of weights $\beta \in \mathbb{R}^{K}$. Thus, the network under consideration is defined by the function:
\begin{equation*}
\Omega_{\theta}(X) \equiv \beta_1 h (w_1^{\top} X) + \ldots + \beta_K h(w_K^{\top} X) = \beta^{\top} H(X), 
\end{equation*} 
where $H(X) = (h(w_1^{\top}X), \ldots, h(w_K^{\top} X))^{\top}$. The neural network is parameterized by $\theta \equiv (\beta^{\top}, w_1^{\top}, \ldots, w_{k}^{\top})^{\top}$. Under this model, the distribution of $Y|X$ is a GLM model with covariates $H(X)$. 

\subsubsection{Statistician's Objective Function} We will endow the statistician with the loss function given by the negative of the conditional log-likelihood for the model in \eqref{eqn:NN}.  

\subsubsection{Nature's Uncertainty Set} We allow nature to introduce additional noise to the statistician's model. We do this in two steps. First, we allow nature to distort the distribution of $X$ using a multiplicative noise denoted as $\xi({1}) \in \mathbb{R}^d$. This is exactly analogous to what we did in the GLM model, where nature was allowed to pick a distribution for the covariates of the form $(X \odot \xi(1))$. Using the jargon of neural networks, we allow nature to contaminate the \emph{input layer} with independent and multiplicative noise.  Second, we also allow nature to contaminate \emph{each of the hidden units} with multiplicative noise $\xi(2) \in \mathbb{R}^{K}$. That is, nature is also allowed to pick a vector $\xi(2) = (\xi(2)_1, \ldots, \xi(2)_{K})^\top$, independently of $\xi(1) \in \mathbb{R}^d$, to distort the each of the $K$ units in the hidden layer as
\[ H(X) \odot \xi(2) \equiv (h(w_1^{\top}X)\xi(2)_1, \ldots, h(w_K^{\top}X)\xi(2)_K)^{\top}. \]

Our choice of a one-hidden neural network was simply for expositional simplicity, but the analysis would be the same with a feed-forward neural network with $L$ hidden layers.  

\subsubsection{Minimax Solution} The minimax solution of the DRO game is given by
\begin{equation} \label{eqn:DRO_NN}
\inf_{\theta} \sup_{\QQ} \mathbb{E}_{\mathbb{Q}} \left[ -\ln f(Y| H(X \odot \xi_1) \odot \xi_2, \beta, \phi )\right],    
\end{equation}
where $\QQ$ now refers to the joint distribution of $(X,Y,\xi(1), \xi(2))$ and $f(Y|X,\beta,\phi)$ is the GLM density defined in (\ref{eqn:GLM}). We continue working with the assumption that $\xi \equiv (\xi(1)^{\top},\xi(2)^{\top})^{\top}$ has independent marginals and that it is independent of $(X,Y)$.

We would like to solve for the worst-case distributions of the random vectors $\xi(1)$ and $\xi(2)$,  assuming that both of these satisfy the restrictions analogous to \eqref{eqn:natures_choice}. The solution for the distribution of $\xi(2)$ can be obtained as a corollary to Theorem \ref{thm:glm}, as it suffices to define
\[ \tilde{X} \equiv H(X \odot \xi(1)),
\]
and view \eqref{eqn:DRO_NN} as the DRO problem in a linear regression model, in which the data is $(\tilde{X},Y)$ and $\xi(2) \in \mathbb{R}^K$ is simply the multiplicative noise that transforms the covariates into $(\tilde{X} \odot \xi(2))$. 

The worst-case choice of $\xi(1)$, the multiplicative error for the inputs, is more difficult to characterize and we were not able to find general results for it. Below, we provide a heuristic argument suggesting that dropout noise might approximate the worst-case choice when the output layer is a Gaussian linear model. Let $\xi(1)_j$ denote the $j$-th coordinate of $\xi(1)$. Suppose that the distribution of this random variable places most of its mass on the interval $[1-\epsilon, 1+\epsilon]$.\footnote{This is compatible with dropout noise for which $\delta$ is very close to zero.} This allows us to `linearize’ the output of each of the hidden units around the output corresponding to unperturbed inputs as
\begin{eqnarray*}
h(w_k^{\top} (X \odot \xi(1))) &=& h(w_k^{\top} (X \odot (\xi(1)-\mathbf{1})) + w_k^{\top}X )  \\
&\approx& h(w_k^{\top}X) +  \left(  \dot{h}(w_k^{\top} X) \cdot  (w_k  \odot X) ^{\top}(\xi(1)-\mathbf{1})  \right).
\end{eqnarray*}
In the notation above, $\mathbf{1}$ denotes the $d$-dimensional vector of ones. For the sake of exposition, ignore the approximation error in the linearization above. If we fix $(X,Y,\xi(2))$, then the worst-case choice for the distribution of $\xi(1)$, denoted by $\QQ(1)$, maximizes
\begin{equation*}
  \EE_{\QQ(1)} \left[  \left(  \sum_{k=1}^{K} \beta_k \cdot \xi(2)_k \cdot \left[  h(w_k^{\top}X) +  \left(  \dot{h}(w_k^{\top} X) \cdot  \sum_{j=1}^{d} w_{k,j} \cdot X_{j} \cdot (\xi(1)_j-1)  \right)  \right]   \right)^2 \right]
\end{equation*}
among all distributions with independent marginals for which  $\mathbb{E}_{\QQ(1)}[ \xi(1)_j ]=1$ for all $j=1, \ldots, d$. Algebra shows that such maximization problem is equivalent to maximizing
\begin{equation} \label{eq:Q1}
  \EE_{\QQ(1)} \left[  \left(  \sum_{k=1}^{K} \beta_k \cdot \xi(2)_k \cdot  \dot{h}(w_k^{\top} X) \cdot  \left[ \sum_{j=1}^{d} w_{k,j} \cdot X_{j} \cdot (\xi(1)_j-1)   \right]   \right)^2 \right],
\end{equation}
which in turn can be written as
\begin{equation*}
  \EE_{\QQ(1)} \left[  \left( a^{\top} (\xi(1)-\mathbf{1})   \right)^2   \right]
\end{equation*}
for an appropriate choice of a vector $a \in \mathbb{R}^d$ that depends only on $(\beta,\xi(2),h,\dot{h},w,X)$. Proposition \ref{prop:exp} in Appendix \ref{subsection:proofTheorem} shows that the solution to this problem is dropout noise. 


%% file: 8Conclusions.tex

In this paper we studied \emph{dropout training}, an increasingly popular estimation method in machine learning. Dropout training is a fundamental part of the modern machine learning techniques for training very deep networks \citep{DLbook:2016}. 

Our main result (Theorem \ref{thm:glm}) established a novel decision-theoretic foundation for the use of dropout training. We showed that this method, when applied to Generalized Linear Models, can be viewed as the minimax solution to an adversarial two-player, zero-sum game between a statistician and nature. The framework used in this paper is known in the stochastic optimization literature \citep{Shapiro:2014} as a Distributionally Robust Optimization (DRO) problem. 

Our minimaxity result showed, by construction, that dropout training indeed provides out-of-sample performance guarantees for distributions that arise from multiplicative perturbations of the in-sample data. Our result thus justified explicitly the ability of dropout training to enhance the out-of-sample performance, which is one of the reasons often invoked to promote the dropout method. 

In addition to our theoretical result, we also suggested a new strategy to select the dropout probability and a new stochastic optimization implementation of dropout training. For the latter, we borrowed ideas from the Multi-level Monte Carlo literature---in particular from the work of \citep{ref:blanchet2019unbiased}---to suggest an unbiased dropout training routine that is easily parallelizable and that has a smaller computational cost compared to naive dropout training methods when the number of features is large (Theorem \ref{thm:unbias}). Crucially, we showed that under some regularity conditions our estimator has finite variance (which means there are also theoretical, and not just practical, gains from parallelization). 

We also discussed the extent to which our theoretical results extended to Neural Networks (in particular, to the universal approximators in \citep{Hornik:1989} consisting of a single-hidden layer and a squashing activation function). Our results showed that Theorem \ref{thm:glm} can be used to establish the optimality of dropout training to estimate the parameters of the last hidden layer in general feed-forward neural networks,  where the output layer takes the form of a Generalized Linear Model. We hope that our analysis serves as a foundation to understand the benefits of dropout training in Neural Networks.

%% file: Appendix.tex
\subsection{Proof of Proposition \ref{prop:limit}} 
\label{subsection:limit}

Algebra shows that the dropout estimator of $\beta$ maximizes
\begin{align*}
Q_n(\beta) 
& \equiv \frac{1}{n}\sum_{i=1}^ny_i(\beta^\top x_i)-\mathbb{E}_{\delta_n}[\Psi(\beta^\top(x_i\odot\xi))].
\end{align*}
In a slight abuse of notation, let $\xi_\delta$ denote a realization of dropout noise parameterized by $\delta$. Then it is possible to re-write the objective function as a weighted average of the functions 
\[
Q_{n,\xi_{\delta_n}}(\beta) \equiv \frac{1}{n}\sum_{i=1}^ny_i(\beta^\top x_i) -\Psi(\beta^\top(x_i\odot\xi_{\delta_n})).
\]
It will be convenient then to define the limiting objective function to be  
\[
Q(\beta) \equiv \mathbb{E}_{P^\star}[Y(\beta^\top X)]-\mathbb{E}_{P^\star}[\mathbb{E}_{\delta}[\Psi(\beta^\top (X\odot\xi))]],
\]
which, by Assumptions 1 and 2, is finite and strictly concave. The population objective function is then the average (over dropout noise) of
\[
Q_{\xi_{\delta}}(\beta) \equiv  \mathbb{E}_{P^\star}[Y(\beta^\top X)]-\mathbb{E}_{P^\star}[\Psi(\beta^\top (X\odot\xi_{\delta}))].
\] 
It is straightforward to show that  $\beta^*(\delta)$ in (\ref{eqn:prob_limit}) denote the unique maximizer of $Q(\beta)$.\\

\begin{proof}
The proof follows from standard arguments in the theory of extremum estimators. In particular, it suffices to verify the conditions of Theorem 2.7 in \cite{newey1994large}. 

Condition i) in \cite{newey1994large} requires $Q(\beta)$ to be uniquely maximized at $\beta^*(\delta)$. This holds because Assumptions 1 and 2 imply that $Q(\beta)$ is strictly concave. 

Condition ii) in \cite{newey1994large} requires $\beta^*(\delta)$ to be an element in the interior of a strictly convex set, which holds because in the GLM models under consideration the parameter space is $\mathbb{R}^{d}$. Furthermore, $Q_n(\beta)$ is trivially concave by Assumption 1.

Condition iii) requires $Q_n(\beta)$ to converge in probability to $Q(\beta)$ for every $\beta$. For this purpose, it suffices to show that  $Q_{n,\xi_{\delta_n}}(\beta)$ converges in probability to $Q_{\xi_{\delta}}(\beta)$ for each fixed $\beta$, and for a sequence $\xi_{\delta_n}$ and $\xi_{\delta}$ that have zeros and non-zeros in exactly the same entries. Assumptions 1 and 2 imply $\mathbb{E}_{P^*} [ Y(\beta^\top X) ]<\infty$ for all $\beta$. Thus, using the 
Law of Large Numbers for i.i.d sequences  
\[ \frac{1}{n}\sum_{i=1}^nY_i(\beta^\top X_i) \overset{p}{\rightarrow} \mathbb{E}_{P^*} [ Y(\beta^\top X) ].   \]
Finally, Assumptions 1 and 2 imply that the triangular array
\[
Z_{n,i} = \Psi(\beta^\top(X_i\odot\xi_{\delta_n})), \quad 1 \leq i \leq n,
\]
satisfies the conditions for the Law of Large Numbers for triangular arrays (Theorem 2.2.11 in~\cite{ref:durret2010probability}), and consequently
\[ \frac{1}{n}\sum_{i=1}^n  \Psi(\beta^\top(X_i\odot\xi_{\delta_n})) \overset{p}{\rightarrow}  \mathbb{E}_{P^*} \left[ \Psi(\beta^\top(X \odot\xi_{\delta}))  \right]. \]
This completes the proof.
\end{proof}

\subsection{Proof of Theorem \ref{thm:glm}} \label{subsection:proofTheorem}

The proof of Theorem~\ref{thm:glm} relies on the following two preparatory results.

\begin{lemma}[Extremal expectation of a univariate convex function] \label{lemma:convex}
For any $-\infty<a<b<+\infty$, let $\zeta$ be a random variable in $[a,b]$ with mean $\mu\in[a,b]$. For any function $f:[a,b]\to\mathbb{R}$ convex and continuous, the distribution of $\zeta$ that maximizes $\EE[f(\zeta)]$ among all distributions over $[a,b]$ with a given mean $\mu \in [a,b]$ is a scaled and shifted Bernoulli distribution, i.e.,
\be \label{eq:zeta-opt}
    \zeta = \begin{cases}
    a & \text{ with probability }  (b-\mu)/(b-a), \\ b & \text{ with probability }  (\mu-a)/(b-a).
    \end{cases}
\ee
\end{lemma}
\begin{proof}
Let $Q^*$ denote the probability measure induced by the random variable in \eqref{eq:zeta-opt}. 
By definition
\[
\EE_{Q^*}[f(\zeta)] = \frac{b-\mu}{b-a}f(a) + \frac{\mu-a}{b-a}f(b).
\]
Suppose first that $\mu=a$. In this case, Jensen's inequality implies that for any other probability measure $Q$ over  $[a,b]$ with mean $\mu=a$, 
\[ \EE_{Q}[f(\zeta)] \leq f(\EE_{Q}[\zeta]) = f(a)= \EE_{Q^*}[f(\zeta)].\]  
An analogous result holds if $\mu=b$.  

Consider then the case in which $\mu\in(a,b)$. For an arbitrary probability measure $Q$ over $[a,b]$ with mean $\mu \in (a, b)$, we have
\[
\int_{[a,b]} f(\zeta) \mathrm{d}Q = \int_{[a,b]} f\big(a\frac{b-\zeta}{b-a}+b\frac{\zeta-a}{b-a} \big) \mathrm{d}Q \leq \int_{[a,b]} \big(\frac{b-\zeta}{b-a}f(a)+\frac{\zeta-a}{b-a}f(b) \big)\mathrm{d}Q,
\]
where the inequality follows from the convexity of $f$. By the linearity of the integral operator and the fact that $\int_{[a,b]}\zeta\mathrm{d}Q=\mu$, we find
\[
\int_{[a,b]} f(\zeta) \mathrm{d}Q \leq \frac{b-\mu}{b-a}f(a)+\frac{\mu-a}{b-a}f(b).
\]
Because the probability measure $Q$ was chosen arbitrarily, this implies that the distribution of~$\zeta$ in~\eqref{eq:zeta-opt} maximizes the expectation of $f(\zeta)$.
\end{proof}

\begin{proposition} \label{prop:exp}
Fix a vector of tuning parameters $\delta \in (0, 1)^d$. Let $\mathcal{Q}_j(\delta_j)$ be defined as in \eqref{eqn:Qj}. Suppose that $A$ is a convex and continuous function on  $\R$.  
For any $\theta\in\mathbb{R}^d$, we have
\[
\sup\left\{ \EE_{\QQ_1 \otimes \ldots \otimes \QQ_d}[A(\theta^\top \xi)]: \QQ_j \in \mc Q_j(\delta_j)\right\}= \EE_{ \QQ\opt_1 \otimes \ldots \otimes \QQ\opt_d} [ A(\theta^\top\xi)],
\]
where $\QQ\opt_j$ is a scaled Bernoulli distribution of the form $\QQ_j\opt= (1-\delta_j)^{-1} \times\textrm{Bernoulli}((1-\delta_j))$ for each $j=1,\ldots,d$.
\end{proposition}


\begin{proof}
First note that $\QQ_j\opt \in \mc Q_j(\delta_j)$ for each $j$, and thus $\QQ_1\opt \otimes \ldots \otimes \QQ_d\opt$ is a feasible solution to the maximization problem. It suffices to show that for any set of feasible measures $\QQ_j\in\mathcal{Q}_j(\delta_j),j=1,\ldots,d$, we have
\[
\EE_{\QQ_1 \otimes \ldots \otimes \QQ_d}[A(\theta^\top \xi)]\leq\EE_{\QQ_1\opt \otimes \ldots \otimes \QQ_d\opt}[A(\theta^\top \xi)].
\]
Towards this end, pick any $k \in \{1, \ldots, d\}$. By Fubini's theorem, we can write
\[
    \EE_{\QQ_1 \otimes \ldots \otimes \QQ_d}[A(\theta^\top \xi)]=\EE_{\QQ_1 \otimes \ldots \otimes \QQ_{k-1} \otimes \QQ_{k+1}\otimes \ldots \otimes \QQ_d} \EE_{\QQ_k}[A(\theta^\top \xi)].
\]
For any fixed value $(\xi_1, \ldots, \xi_{k-1}, \xi_{k+1}, \ldots, \xi_d)$ the function $\xi_k \mapsto A(\sum_{j\neq k}\theta_j\xi_j + \theta_k \xi_k)$ is convex in the variable $\xi_k$ over the interval $[0,(1-\delta_k)^{-1}]$. Thus by Lemma~\ref{lemma:convex}, 
\[
    \EE_{\QQ_k}[A(\sum_{j\neq k} \theta_j \xi_j + \theta_k \xi_k)]\leq\EE_{\QQ_k\opt}[A(\sum_{j\neq k} \theta_j \xi_j + \theta_k \xi_k)] \quad\textrm{for any fixed } (\xi_1, \ldots, \xi_{k-1}, \xi_{k+1}, \ldots, \xi_d).
\]
Thus by the monotonicity of the expectation operator,
\[
     \EE_{\QQ_1 \otimes \ldots \otimes \QQ_d}[A(\theta^\top \xi)] \leq \EE_{\QQ_1 \otimes \ldots \otimes \QQ_{k-1} \otimes \QQ_{k+1}\otimes \ldots \otimes \QQ_d} \EE_{\QQ_k\opt}[A(\theta^\top \xi)] = \EE_{\QQ_1 \otimes \ldots \otimes \QQ_{k-1} \otimes \QQ_k\opt \otimes \QQ_{k+1}\otimes \ldots \otimes \QQ_d} [A(\theta^\top \xi)] .
\]
By cycling through all possible values of $k \in \{1, \ldots, d\}$ we conclude that
\[
\EE_{\QQ_1 \otimes \ldots \otimes \QQ_d}[A(\theta^\top \xi)] \leq \EE_{\QQ_1\opt \otimes \ldots \otimes \QQ_d\opt} [ A(\theta^\top\xi)].
\]
Therefore, the postulated claim holds.
\end{proof}

We are now ready to prove Theorem~\ref{thm:glm}.
\begin{proof}
Note that for $\QQ\in\mathcal{U}(\mathbb{Q}_0,\delta)$, Assumption \ref{asn:reg} implies $\EE_{\QQ}[\ell(X\odot\xi,Y,
    \theta)]$ is finite for any $\theta \in \Theta$ and any scalar $\delta \in [0,1)$. Therefore, from Fubini's theorem and the definition of loss function:
\begin{align*}
    \EE_{\QQ}[\ell(X\odot\xi,Y,
    \theta)]&=\mathbb{E}_{\mathbb{Q}_0}  \left[ \EE_{\QQ_1 \otimes \ldots \otimes \QQ_d}[\ell(X \odot \xi,  Y,
    \theta)] \right] \\
    &=\mathbb{E}_{\mathbb{Q}_0}  \left[ \EE_{\QQ_1 \otimes \ldots \otimes \QQ_d}[-\ln h(Y,\phi) + (\Psi(\beta^\top (X \odot \xi))-Y(\beta^\top (X \odot \xi)) )/a(\phi)] \right] \\
    &=-\mathbb{E}_{\mathbb{Q}_0} \left[  \ln h(Y,\phi) \right] \\
    & +  \mathbb{E}_{\mathbb{Q}_0}  \left[ \EE_{\QQ_1 \otimes \ldots \otimes \QQ_d}[(\Psi(\beta^\top (X \odot \xi) )-Y(\beta^\top (X \odot \xi)) )/a(\phi)] \right]. 
\end{align*}
Algebra shows that for any $\beta$, $X$ and $\xi$:
\[ \beta^\top (X \odot \xi) = (\beta \odot X)^\top \xi. \]
Thus, we can fix the values of $(X,Y, \theta)$ and define the function
\[
    A_{(X,Y,\theta)} ( (\beta \odot X)^\top \xi ) \equiv (\Psi(\beta^\top ( X \odot\xi))- Y \beta^\top (X \odot\xi))/a(\phi).
\]
Note that $A_{(X,Y,\theta)}$ satisfies the condition of Proposition~\ref{prop:exp}. Therefore
\[
\sup\left\{ \EE_{\QQ_1 \otimes \ldots \otimes \QQ_d}[A_{(X,Y,\theta)} ( (\beta \odot X)^\top \xi )]: \QQ_j \in \mc Q_j(\delta_j)\right\} = \mathbb{E}_{\QQ_1\opt \otimes \ldots \otimes \QQ_d\opt}[A_{(X,Y,\theta)} ( (\beta \odot X)^\top \xi )],
\]
for any $(X,Y,\theta)$, which completes the proof.
\end{proof}

    

\subsection{Proof of Proposition \ref{prop:loss}}
\label{subsection:loss}

\begin{proof}
We write   $\sqrt{n}(\mathcal{L}_n(\widehat{\beta} (\delta_n),\widehat{\phi},\delta_n)-\mathcal{L}(\beta^*,\phi^*))$ as the sum of the following three terms
\begin{subequations}
\begin{equation}\label{eqn:Aux3} 
\sqrt{n}\left( \mathcal{L}_n(\widehat{\beta}(\delta_n), \widehat{\phi},\delta_n)-\mathcal{L}_n(\beta^*,\widehat{\phi}, \delta_n) \right),
\end{equation}
\begin{equation}  \label{eqn:Aux4}
\sqrt{n}\left( \mathcal{L}_n(\beta^*, \widehat{\phi}, \delta_n)-\mathcal{L}_n(\beta^*,\widehat{\phi}, 0) \right),
\end{equation}
\begin{equation}  \label{eqn:Aux5}
\sqrt{n}\left( \mathcal{L}_n(\beta^*,\widehat{\phi},0)-\mathcal{L}(\beta^*,\phi^*, 0) \right).
\end{equation}
\end{subequations}
The last term converges in distribution to a normal random variable, so we only need to analyze \eqref{eqn:Aux3} and \eqref{eqn:Aux4}. 

By Assumptions 1 and 2 the term in  \eqref{eqn:Aux3} admits an exact second-order Taylor expansion around $\beta^*$ for every $\phi$ and $\delta$. We argue that because of this, the term in question if $o_{p}(1)$. First, using the same arguments as in Theorem 3.1 in \cite{newey1994large} we can show that for any sequence $\delta_n=c/\sqrt{n}$
\[
\sqrt{n}(\widehat{\beta}(\delta_n)-\beta^\star) \overset{d}{\rightarrow}  \Sigma(\beta^*)^{-1}\mathcal{N}_d(-c\tilde\mu,a(\phi^*)\Sigma(\beta^*)),
\]
where
\[ \Sigma (\beta)  \equiv \mathbb{E}_{P^*} [\ddot{\Psi}(X^{\top} \beta ) X X^{\top} ], \]
and
\[
\tilde\mu \equiv \left( \sum_{\xi\in \mathcal{A}}\mathbb{E}_{P^\star}[\dot{\Psi}((X\odot\xi)^\top \beta^\star )(X\odot\xi)] \right)-(d-1)\mathbb{E}_{P^\star}[YX]+ \Sigma(\beta^*)\beta^*.
\]
The set $\mathcal{A}$ above is defined as  $\{\xi\in\{0,1\}^d:\textrm{exactly one entry of }\xi\textrm{ is zero}\}$. The argument is essentially the same as in every proof of asymptotic normality for extremum (or $M$-estimators), with the only difference being that, because of the dropout noise, the score term is asymptotically normal with a nonzero mean.  In fact, 
\[ \nabla_{\beta} \mathcal{L}_n(\beta,\widehat{\phi}, \delta_n) \equiv \nabla_{\beta} \mathcal{L}_n(\beta,\widehat{\phi}, 0) + \frac{1}{a(\widehat{\phi})}\left( \frac{1}{n} \sum_{i=1}^{n} \mathbb{E}_{\delta_n} [  (X_i \odot \xi) \dot{\Psi} (\beta^{\top} (X_i \odot \xi)) ] - X_i \dot{\Psi}(X_i^{\top} \beta) \right) , \]
where
\[ \nabla_{\beta} \mathcal{L}_n(\beta,\widehat{\phi}, 0) \equiv -\frac{1}{a(\widehat{\phi})} \frac{1}{n} \sum_{i=1}^{n} X_i (Y_i - \dot{\Psi}(X_i^{\top} \beta)). \]
Recognizing the term $\nabla_{\beta} \mathcal{L}_n(\beta,\widehat{\phi}, 0)$ as the negative of the score function in the GLM model and doing some algebra, it is possible to show that $\nabla_{\beta} \mathcal{L}_n(\beta,\widehat{\phi}, \delta_n)$ is $o_{p}(1)$. 

For the term in \eqref{eqn:Aux4}, note first that it is nonnegative. Also: $\mathcal{L}_n(\beta^*,\widehat{\phi},\delta_n)-\mathcal{L}_n(\beta^*,\widehat{\phi}, 0)$ equals
\[ \frac{1}{a(\widehat{\phi})} \left( \frac{1}{\sqrt{n}} \sum_{i=1}^{n} \mathbb{E}_{\delta_n}[\Psi( (X_i \odot \xi)^{\top} \beta^* ) ] - \Psi( X_i^{\top} \beta^* ) \right).    \]
The term in parenthesis has finite mean equal to
\begin{equation}
    \Delta_n  \equiv  \mathbb{E}_{P^*} \mathbb{E}_{\delta_n}[\Psi( (X \odot \xi)^{\top} \beta^* ) ] - \mathbb{E}_{P^*}[ \Psi( X^{\top} \beta^* ) ]   .
\end{equation}
It can be shown---by verifying the conditions for the Law of Large Numbers for triangular arrays (Theorem 2.2.11 in \cite{ref:durret2010probability})---that
\[\sqrt{n}(\mathcal{L}_n(\beta^*,\delta_n)-\mathcal{L}_n(\beta^*,0) - a(\widehat{\phi})^{-1} \Delta_n ) \overset{p}{\rightarrow} 0. \]
Moreover, Assumptions 1 and 2 imply 
\[ \sqrt{n} \Delta_n \overset{p}{\rightarrow} \Delta, \]
where 
\[ \Delta \equiv 
c\left(\sum_{\xi\in \mathcal{A}}\mathbb{E}_{P^\star}[\Psi((X\odot\xi)^{\top} \beta^\star)  ]-d\mathbb{E}_{P^\star}[\Psi(X ^\top \beta^\star )]+\mathbb{E}_{P^\star}[\dot{\Psi}(X^{\top} \beta^*) X^\top \beta^\star ]\right).
\]  

\noindent This gives the desired result. 

\end{proof}

\subsection{Proof of Theorem \ref{thm:unbias}} \label{subsection:Proof_Bias}

\begin{proof}
By definition
\[
    Z (K^*_l) = \frac{\bar\Delta_{{K_l^*}}}{ r (1-r)^{m^*_l} } +\theta_{l,{m_0}},
\]
where $K^*_l$ is a discrete random variable with probability mass function:
\[ p(K^*_l) = r(1-r)^{K^*_l-m_0}, \]
and supported on the integers larger than $m_0$. 

We first show that the estimator $Z (K^*_l)$ is unbiased (as we average over both $K^*_l$ and $\xi^k_{i}$). Algebra shows that
\begin{eqnarray*}
\mathbb{E}[ Z(K^*_l) ] &=& \sum_{K=m_0}^{\infty} \mathbb{E}[ Z(K^*_l)  | K^*_l = K  ] p(K)  \\
&=& \sum_{K=m_0}^{\infty} \mathbb{E} \left[ \frac{\bar{\Delta}_{K^*_l}}{p(K^*_l)} + \theta_{l,m_0}  \Bigg | K^*_l = K  \right] p(K) \\
&=& \sum_{K=m_0}^{\infty} \mathbb{E} \left[ \frac{\bar{\Delta}_{K}}{p(K)} + \theta_{l,m_0}  \Bigg | K^*_l = K  \right] p(K) \\
&=& 
   \left( \sum_{K=m_0}^{\infty} \mathbb{E} \left[ \widehat\theta^\star_{n}(2^{K+1})-\frac{1}{2}(\widehat\theta^{O}_{n}(2^K)+\widehat\theta^{E}_{n}(2^K))  \right] \right) + \mathbb{E}[\theta_{l,m_0}] \\ 
&=& -\frac{1}{2} \left( \mathbb{E}[\widehat\theta^{O}_{n}(2^{m_0})] + \mathbb{E}[\widehat\theta^{E}_{n}(2^{m_0})] \right)  + \mathbb{E}[\theta_{l,m_0}] + \lim_{K \rightarrow \infty} \mathbb{E}[\widehat\theta^\star_{n}(2^{K+1})].    
\end{eqnarray*}
The expectations in the last line are all finite because $\Theta$ is compact. In addition, since the draws are i.i.d.~and $\theta_{l,m_0}$ is the solution to the problem \eqref{eq:sampopt} when $2^{m_0}$ draws are used we have
\[-\frac{1}{2} \left( \mathbb{E}[\widehat\theta^{O}_{n}(2^m_0)] + \mathbb{E}[\widehat\theta^{E}_{n}(2^m_0)] \right)  + \mathbb{E}[\theta_{l,m_0}] = 0.\]
Moreover, the sequence of random variables
\[ \{ \widehat\theta^\star_n(2^{K+1}) \} \]
is uniformly integrable, because $\Theta$ is a compact subset of a finite-dimensional Euclidean space. Finally, we know that
\[ \widehat\theta^\star_{n}(2^{K+1}) \overset{p}{\rightarrow} \theta\opt_n\]
as $K \rightarrow \infty$. The uniform integrability of the sequence of estimators then implies
\[  \lim_{K\to\infty}\mathbb{E}[\widehat\theta^\star_{n}(2^{K+1})] = \mathbb{E}\left[\lim_{K \rightarrow \infty} \widehat\theta^\star_{n}(2^{K+1}) \right] = \theta\opt_n, \]
see Theorem 6.2 in \cite{Dasgupta08}. We conclude that
\[ \mathbb{E}[ Z(K^*_l) ] = \lim_{K \rightarrow \infty} \mathbb{E}[\widehat{\theta}^\star_{n}(2^{K+1})] = \theta\opt_n. \]

Now we show that the expected computational cost of $Z(K^*_l)$ is finite. In order to compute $Z(K)$ for a given $K$ we need $n \cdot 2^{K+1}$ random draws. Thus, the expected computational cost of $Z(K^*_l)$ is
\begin{eqnarray*}
\sum_{K=m_0}^{\infty} n 2^{K+1} r (1-r)^{K-m_0} &=& n \cdot (2^{m_0+1}) \cdot r   \sum_{K=m_0}^{\infty} 2^{K-m_0} (1-r)^{K-m_0} \\
&=& n \cdot (2^{m_0+1}) \cdot r   \sum_{K=m_0}^{\infty} (2(1-r))^{K-m_0}. 
\end{eqnarray*}
The term above converges to
\[  \frac{n \cdot (2^{m_0+1}) \cdot r }{1-2(1-r)} = \frac{n \cdot (2^{m_0+1}) \cdot r }{2r-1} \]
provided that $2(1-r)<1$, which holds because we have chosen $r>1/2$. 

For the proof on finite variance, we intend to show that 
\begin{equation}\label{eq:finitevar1}
\EE\left[\bar\Delta_{K}^\top \bar\Delta_{K}\right] = O(2^{-2K})
\end{equation}
as $K\to\infty$. Equation \eqref{eq:finitevar1} guarantees that every processor generates an estimator $Z(K^*_l)$ with finite variance. Since $K_l^*$ is a discrete random variable with probability mass function
\[ p(K^*_l) = r(1-r)^{K^*-m_0}, \]
and
\begin{eqnarray*}
\EE[Z(K_l^*)^\top Z(K_l^*)] & =& \sum_{K=m_0}^\infty \EE\left[Z(K_l^*)^\top Z(K_l^*)|K_l^* = K \right] p(K)\\
&=& \sum_{K=m_0}^\infty \EE\left[\left(\frac{\bar\Delta_{K}}{p(K)}+\theta_{l,m_0}\right)^\top \left(\frac{\bar\Delta_{K}}{p(K)}+\theta_{l,m_0}\right)\right]p(K)\\
&\leq& 2\left(\sum_{K=m_0}^\infty \EE\left[\frac{\bar\Delta_K^\top\bar\Delta_K}{p(K)^2}\right]p(K) + \sum_{K=m_0}^\infty \EE\left[\theta_{l,m_0}^\top\theta_{l,m_0}\right]p(K)\right)\\
&\leq& C \left(\sum_{K=m_0}^\infty \frac{2^{-2K}}{p(K)}+ \sup_{\theta\in\Theta}\|\theta\|_2^2p(K)\right)\\
&\leq& C \left(\sum_{K=m_0}^\infty \frac{1}{2^{2 m_0} 2^{2(K-m_0)}p(K)}+ \sup_{\theta\in\Theta}\|\theta\|_2^2p(K)\right)\\
&\leq& C_1\left(\sum_{K=m_0}^\infty \frac{1}{r4^{m_0}}\frac{1}{(4(1-r))^{K-{m_0}}}\right)+C_2.
\end{eqnarray*}
The geometric sum in the last expression is finite because we have assumed that  $r<\frac{3}{4}$.

To show~\eqref{eq:finitevar1}, we do a Taylor expansion of the first-order conditions of the problem  \eqref{eq:sampopt} around $\theta\opt_n$. The Karush-Kuhn-Tucker optimality condition for the level $2^K$ solution $\widehat\theta_n^\star(2^K)$ of the problem in~\eqref{eq:sampopt} implies
\[
0 = \sum_{i=1}^n\left[\frac{1}{2^K}\sum_{k=1}^{2^K}\nabla_\theta\ell(x_i\odot\xi_i^k,y_i,\widehat\theta_n^\star(2^K))\right].
\]
It follows by the Taylor expansion and Assumption 4 that 
\begin{eqnarray}
 0 &=& \sum_{i=1}^n\left[\frac{1}{2^K}\sum_{k=1}^{2^K}\nabla_\theta\ell(x_i\odot\xi_i^k,y_i,\theta_n^\star)\right] + \sum_{i=1}^n\left[\frac{1}{2^K}\sum_{k=1}^{2^K}\nabla_{\theta\theta} \ell(x_i\odot\xi_i^k,y_i,\theta_n^\star)\right]\left(\widehat\theta_n^\star(2^K)-\theta_n^\star\right)\notag\\
 & & + R_{K,\theta}\notag\\
 & = &  \sum_{i=1}^n\left[\frac{1}{2^K}\sum_{k=1}^{2^K}\nabla_\theta\ell(x_i\odot\xi_i^k,y_i,\theta_n^\star)\right] + \sum_{i=1}^n \nabla_{\theta\theta}\EE_{\QQ^\star}\left[\ell(X\odot\xi,Y,\theta_n^\star)|X = x_i, Y= y_i\right]]\left(\widehat\theta_n^\star(2^K)-\theta_n^\star\right)\notag\\
 & & + R_K + R_{K,\theta},\label{eq:reterms}
\end{eqnarray}
where
\[
R_K \equiv \left(\sum_{i=1}^n\left(\frac{1}{2^K}\sum_{k=1}^{2^K}\nabla_{\theta\theta}\ell(x_i\odot\xi_i^k,y_i,\theta_n^\star)-\nabla_{\theta\theta}\EE_{\QQ^\star}\left[\ell(X\odot\xi,Y,\theta_n^\star)|X = x_i, Y= y_i\right]\right)\right)\left(\widehat\theta_n^\star(2^K)-\theta_n^\star\right)
\]
and 
\[
\|R_{K,\theta}\|_2 \leq \sum_{i=1}^n\sup_{\theta\in\Theta,\xi}\left\| \nabla_{\theta\theta\theta} \ell(x_i\odot \xi, y_i,\theta)\right\|_2 \left\|\widehat\theta_n^\star(2^K)-\theta_n^\star\right\|_2^2\leq C_3 \left\|\widehat\theta_n^\star(2^K)-\theta_n^\star\right\|_2^2
\]
by Assumption 4. Thus by Assumption 3, we have
\[
\EE[R_{K,\theta}^\top R_{K,\theta}] = O(2^{-2K})
\]
as $K\to\infty$. Moreover, by the multivariate version of Theorem~2 in~\cite{bahr:1965} which follows from the Cram\'{e}r-Wold theorem, we have that
\[
\EE\left[\left\|\sum_{i=1}^n\left(\frac{1}{2^K}\sum_{k=1}^{2^K}\nabla_{\theta\theta}\ell(x_i\odot\xi_i^k,y_i,\theta_n^\star)- \nabla_{\theta\theta}\EE_{\QQ^\star}\left[\ell(X\odot\xi,Y,\theta_n^\star)|X = x_i, Y= y_i\right]\right)\right\|_2^4\right]
\]
is $O(2^{-2K})$.

We can express $R_{K}^{\top}R_{K}$ as $ \left\| R_{K} \right\|^2 $. The Cauchy-Schwarz inequality implies
\begin{align*}
&\EE[R_K^\top R_K]\\
\leq& \EE\left[\left\|\sum_{i=1}^n\left(\frac{1}{2^K}\sum_{k=1}^{2^K}\nabla_{\theta\theta}\ell(x_i\odot\xi_i^k,y_i,\theta_n^\star)- \nabla_{\theta\theta}\EE_{\QQ^\star}\left[\ell(X\odot\xi,Y,\theta_n^\star)|X = x_i, Y= y_i\right]\right)\right\|_2^2  \cdot \right.\\
& \left. \left\|\widehat\theta_n^\star(2^K)-\theta_n^\star\right\|_2^2\right].\\
\end{align*}
By H\"{o}lder's inequality we have
\begin{align*}
&\EE[R_K^\top R_K] \\
\leq& \EE\left[\left\|\sum_{i=1}^n\left(\frac{1}{2^K}\sum_{k=1}^{2^K}\nabla_{\theta\theta}\ell(x_i\odot\xi_i^k,y_i,\theta_n^\star)- \nabla_{\theta\theta}\EE_{\QQ^\star}\left[\ell(X\odot\xi,Y,\theta_n^\star)|X = x_i, Y= y_i\right]\right)\right\|_2^4\right]^{\frac{1}{2}} \times \\
& \hspace{3cm} \EE\left[\left\|\widehat\theta_n^\star(2^K)-\theta_n^\star\right\|_2^4\right]^{\frac{1}{2}}\\
\leq& O(2^{-2K}).
\end{align*}
Finally, consider the solutions $\widehat\theta_n^\star(2^{K^\star_l+1}),\widehat\theta_n^O(2^{K_l^\star}),\widehat \theta_n^E(2^{K_l^\star})$ conditional on $K_l^\star = K$. Denote the remainder terms in~\eqref{eq:reterms} corresponding to the level $2^{K+1}$ solution $\widehat\theta_n^\star(2^{K+1})$ as $R^\star_{K+1},R^\star_{K+1,\theta}$. Similarly, denote the remainder terms in~\eqref{eq:reterms} corresponding to the level $2^K$ solution $\widehat \theta_n^O(2^K)$ (and, respectively, $\widehat \theta_n^E(2^K)$) as $R^O_K,R^O_{K,\theta}$ (  $R^E_K,R^E_{K,\theta}$). By the construction of $\widehat \theta_n^O(2^K),\widehat \theta_n^E(2^K)$ using odd and even indices, we have, from~\eqref{eq:reterms}
\begin{align*}
& -\sum_{i=1}^n\nabla_{\theta\theta}\EE_{\QQ^\star}[\ell(X\odot\xi,Y,\theta_n^\star)|X=x_i,Y=y_i]\left(\widehat\theta_n^\star(2^{K+1})-\frac{1}{2}(\widehat\theta_n^O(2^K)+\widehat\theta_n^E(2^K))\right) \\
= & R^\star_{K+1} -\frac{1}{2}(R_K^O+R_K^E) + R^\star_{K+1,\theta}-\frac{1}{2}(R^O_{K,\theta}+R^E_{K,\theta}).
\end{align*} 
By Assumption 4,
\[
\sum_{i=1}^n\nabla_{\theta\theta}\EE_{\QQ^\star}[\ell(X\odot\xi,Y,\theta_n^\star)|X=x_i,Y=y_i] =n\cdot\nabla_{\theta\theta}\EE_{\QQ^\star}[\ell(X\odot\xi,Y,\theta_n^\star)]
\]
is invertible. Thus, we have shown that 
\begin{align*}
\bar\Delta_K &\equiv \widehat\theta_n^\star(2^{K+1})-\frac{1}{2}(\widehat\theta_n^O(2^K)+\widehat\theta_n^E(2^K)) \\
&=   \left( n\cdot\nabla_{\theta\theta}\EE_{\QQ^\star}[\ell(X\odot\xi,Y,\theta_n^\star)] \right)^{-1} \left( R^\star_{K+1} -\frac{1}{2}(R_K^O+R_K^E) + R^\star_{K+1,\theta}-\frac{1}{2}(R^O_{K,\theta}+R^E_{K,\theta}) \right).   
\end{align*}
Since each of the terms on the right-hand side have been shown to be $O(2^{-2K})$, we conclude that  $\EE[\bar\Delta_K^\top\bar\Delta_K]=O(2^{-2K})$.
\end{proof}

\subsection{Dropout Training in Linear Regression} \label{subsection:lr}

\begin{corollary}[Linear regression with $\phi=1$] \label{corol:lr}
For linear regression with $\ell(x, y, \beta) = (\beta^\top x -y)^2$, we have
\[
    \adjustlimits\min_{\beta \in \R^d} \max_{\QQ\in\mathcal{U}(\Pnom_n,\delta) }\EE_{\QQ} \Big[\big(\beta^\top (X \odot \xi)-Y\big)^{2}\Big]=\min_{\beta \in \R^d} \EE_{\QQ\opt} \Big[\big(\beta^\top (X \odot \xi)-Y\big)^{2}\Big],
\]
where $\QQ\opt=\Pnom_n\otimes \QQ_1\opt \otimes \ldots \otimes \QQ_d\opt$
and $\QQ_j\opt=(1-\delta)^{-1}\times\textrm{Bernoulli}(1-\delta)$ for each $j=1,\ldots,d$. Moreover, 
\begin{equation}
    \label{eq:reformlrbeta}
    \min_{\beta \in \R^d} \EE_{\QQ\opt} \Big[\big(\beta^\top (X \odot \xi)-Y\big)^{2}\Big] = \min_{\beta\in\mathbb{R}^d}~\frac{1}{n}\left[(\mathbf{Y}-\mathbf{X}\beta)^\top(\mathbf{Y}-\mathbf{X}\beta)+\frac{\delta}{1-\delta}\beta^\top \mathbf{\Lambda}\beta\right],
\end{equation}
which implies that the dropout training estimator equals 
\[
\widehat{\beta}(\delta)=\left(\mathbf{X}^\top\mathbf{X}+\frac{\delta}{1-\delta} \textrm{diag}(\mathbf{X}^\top\mathbf{X}) \right)^{-1}\mathbf{X}^\top\mathbf{Y}.
\]
Finally, 
if $\mathbb{E}_{P^\star}[XX^{\top}]$ is a diagonal matrix with strictly positive entries then
\[ \widehat{\beta}(\delta) \overset{p}{\rightarrow} (1-\delta) E_{P^*}[XX^{\top}]^{-1} \mathbb{E}[XY] . \]

\end{corollary}

\begin{proof} The first part of the corollary follows directly from \eqref{eq:dro} and \eqref{eq:dro-refor} in our main theorem. The second part of the corollary follows from Proposition \ref{prop:limit}. According to this proposition, the limit of $\widehat{\beta}(\delta)$ is
\[ \beta^*(\delta) = \left( \mathbb{E}_{P^*}[XX^{\top}] + (\delta/1-\delta) \textrm{diag}(\mathbb{E}_{P^*}[XX^{\top}]) \right)^{-1} \mathbb{E}_{P^*}[YX]. \]
Thus, if $\mathbb{E}_{P^*}[XX^{\top}]$ is a diagonal matrix, we obtained the desired limit.

\end{proof}


\subsection{Additional Numerical Results} \label{subsection:addnum}
Here we try to provide some justifications for our choice of parameter. 

\textbf{Learning Rate:} We first fix an all zeros initialization scheme, and vary the learning rate. We summarize the average parameter divergence and 1-standard deviation error for $20$ repetitions of the SGD algorithm in Table~\ref{tab:1}. We can observe the learning rate $0.0001$ shows a clear advantage. 

\textbf{Initialization:} Next we fix the learning rate to be $0.0001$, and consider different initialization schemes.  We note that the mean value (resp., absolute value) of elements in $\beta^\star$ is 
$0.3947$ (resp., $0.6977$). Table~\ref{tab:2} shows the average parameter divergence and the 1-standard deviation from $20$ repetitions of the SGD algorithm. We see that the initialization at origin is a fair choice. 

\begin{table}[htb]
    \centering
    \begin{tabular}{|c|c|c|c|}
    \hline
    Learning rate  & 0.001 & \textbf{0.0001} & 0.00001\\ \hline
    $\|\widehat\beta_{SGD}-\beta^\star\|_\infty$    &  $0.0827\pm0.0133$  & $\mathbf{0.0301\pm0.0025}$ & $0.6702\pm 0.1082$\\
     \hline
     
    \end{tabular}
    \caption{Comparison for different learning rates, with fixed zero initializations.}
    \label{tab:1}
\end{table}

\begin{table}[htb]
    \centering
    \begin{tabular}{|c|c|c|c|}
    \hline
         Initializations & \textbf{all zeros} & all $0.2$'s  & all $1$'s   \\ \hline
         $\|\widehat\beta_{SGD}-\beta^\star\|_\infty$  & $\mathbf{0.0301\pm0.0025}$ &  $0.0317\pm0.0047$ & $0.0614\pm0.0196$ \\
         \hline \hline
          Initializations &  i.i.d~$\mathcal{N}(0,1)$ & i.i.d~$\mathcal{N}(0,10)$ & i.i.d~$\mathcal{N}(0,10^2)$ \\ \hline
         $\|\widehat\beta_{SGD}-\beta^\star\|_\infty$  & $0.0376\pm0.0067$ &  $0.1006\pm0.0469$ & $0.3208\pm0.1432$  \\
         \hline
    \end{tabular} 
    \caption{Comparison for different initialization schemes with fixed learning rate 0.0001.}
    \label{tab:2}
\end{table}

\begin{figure}[h!]
    \centering
    \includegraphics[width=0.6\textwidth]{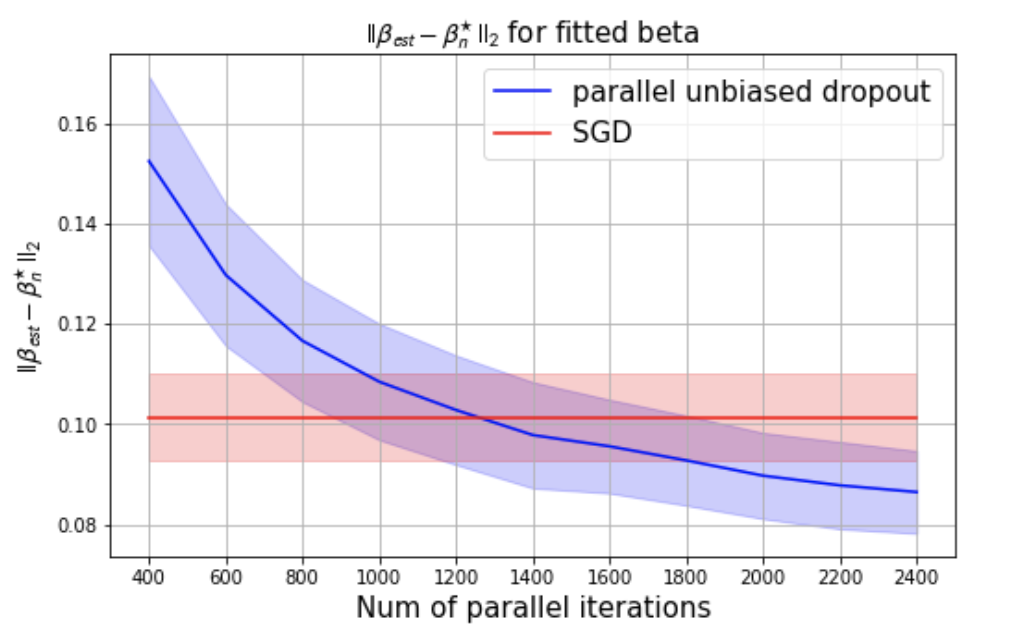}
    \caption{$l_2$ difference for 120s wall-clock time}
    \label{fig:7}
\end{figure}

\begin{figure}[h!]
    \centering
    \includegraphics[width=0.6\textwidth]{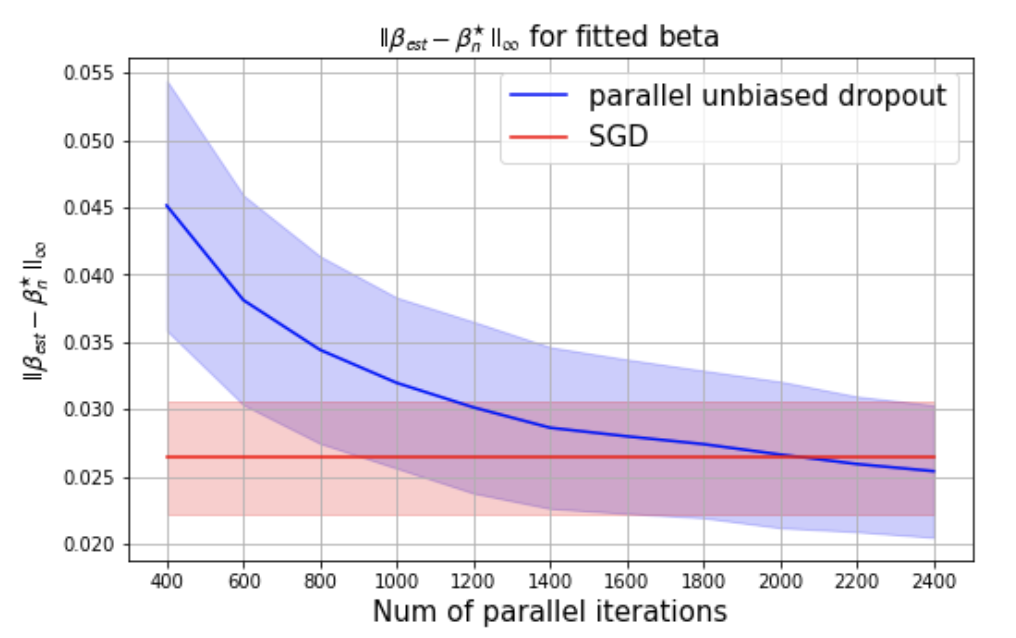}
    \caption{$l_\infty$ difference for 120s wall-clock time}
    \label{fig:8}
\end{figure}

\begin{figure}[h!]
    \centering
    \includegraphics[width=0.6\textwidth]{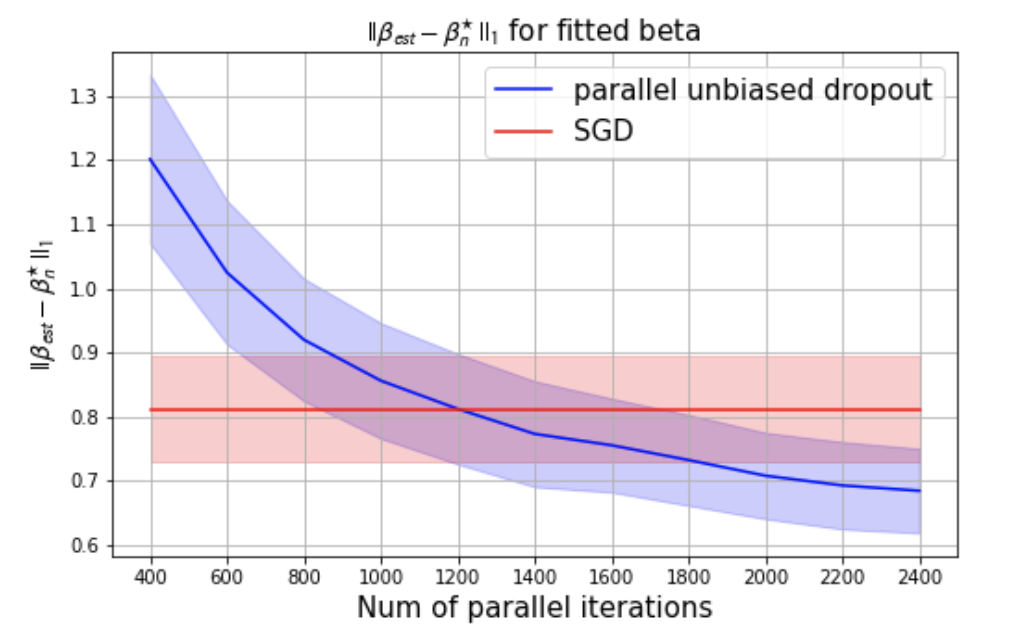}
    \caption{$l_1$ difference for 120s wall-clock time}
    \label{fig:9}
\end{figure}

\begin{figure}[h!]
    \centering
    \includegraphics[width=0.6\textwidth]{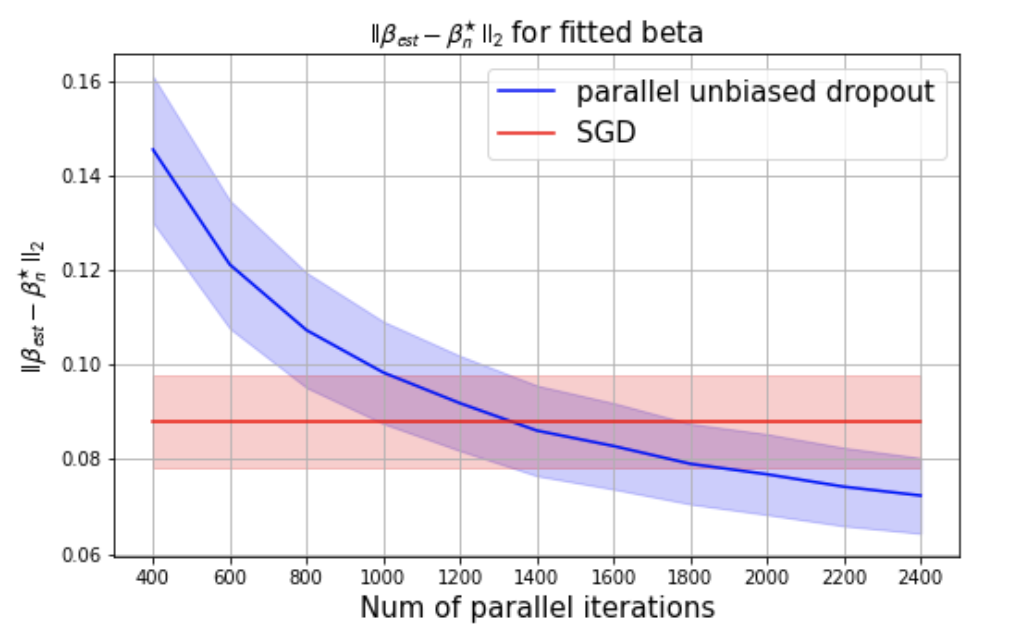}
    \caption{$l_2$ difference for 180s wall-clock time}
    \label{fig:4}
\end{figure}

\begin{figure}[h!]
    \centering
    \includegraphics[width=0.6\textwidth]{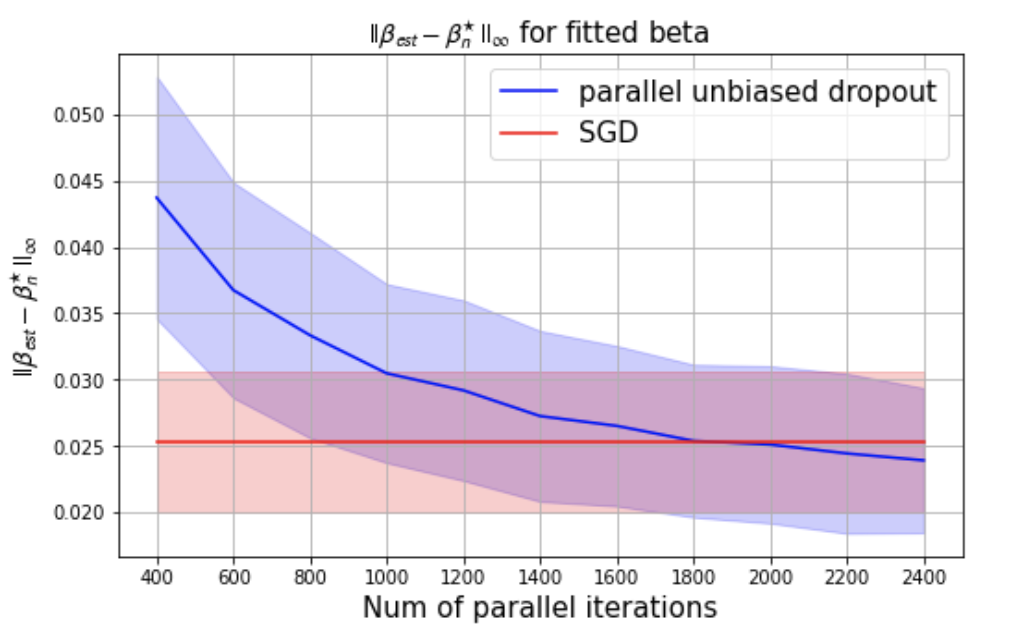}
    \caption{$l_\infty$ difference for 180s wall-clock time}
    \label{fig:5}
\end{figure}

\begin{figure}[h!]
    \centering
    \includegraphics[width=0.6\textwidth]{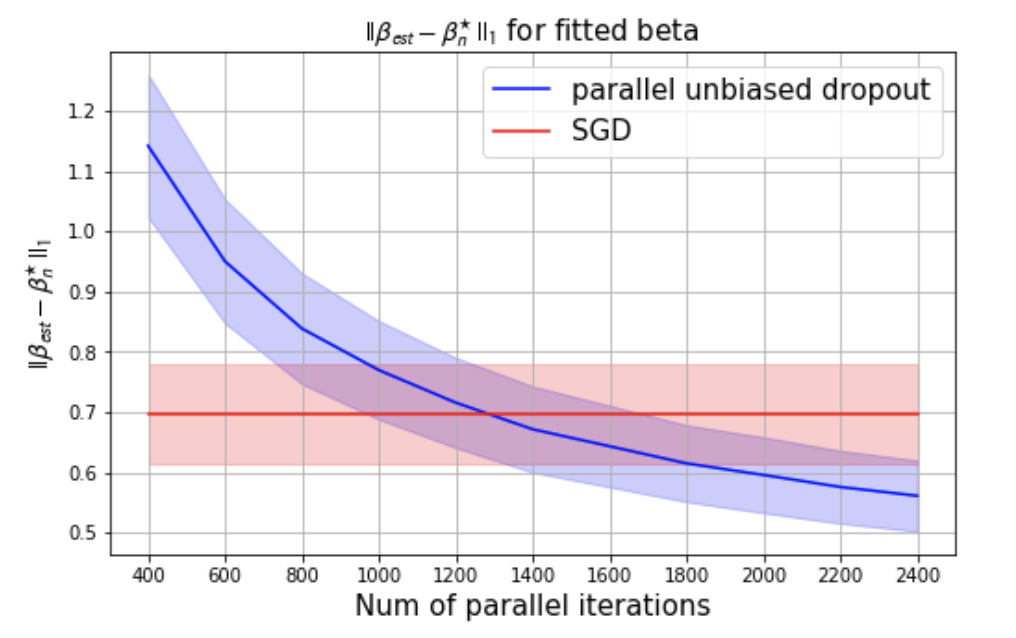}
    \caption{$l_1$ difference for 180s wall-clock time}
    \label{fig:6}
\end{figure}

\textbf{Wall-Clock Time:} We then document the numerical results for 120s/180s wall-clock time, see Figures~\ref{fig:7} -~\ref{fig:9} for the case of 120s and Figures~\ref{fig:4} -~\ref{fig:6} for the case of 180s. We see that the proposed unbiased approach outperforms the standard SGD when the number of parallel iterations reaches above some threshold.




